\documentclass[10pt,twocolumn,letterpaper]{article}

\usepackage[pagenumbers]{iccv} 

\usepackage{colortbl}

\definecolor{iccvblue}{rgb}{0.21,0.49,0.74}
\usepackage[pagebackref,breaklinks,colorlinks,allcolors=iccvblue]{hyperref}

\usepackage{ulem}
\usepackage{multirow}
\usepackage{hhline}
\usepackage{colortbl}
\usepackage{soul}
\usepackage{makecell}

\usepackage{graphicx}

\usepackage{algorithm}
\usepackage{algpseudocode}
\usepackage{multirow}
\newcommand\Algphase[1]{%
\vspace*{-.7\baselineskip}\Statex\hspace*{\dimexpr-\algorithmicindent-2pt\relax}\rule{\columnwidth}{0.4pt}%
\Statex\hspace*{-\algorithmicindent}{#1}%
\vspace*{-.7\baselineskip}\Statex\hspace*{\dimexpr-\algorithmicindent-2pt\relax}\rule{\columnwidth}{0.4pt}%
}

\newcommand\AlgphaseAux[1]{%
\vspace*{-.7\baselineskip}\Statex\hspace*{\dimexpr-\algorithmicindent-2pt\relax}\rule{\columnwidth}{0.4pt}%
}

\def\paperID{12726} 
\def\confName{ICCV}
\def\confYear{2025}

\title{OpenWildlife: Open-Vocabulary Multi-Species Wildlife Detector for Geographically-Diverse Aerial Imagery}

\author{
Muhammed Patel\textsuperscript{1},
Javier Noa Turnes\textsuperscript{1},
Jayden Hsiao\textsuperscript{1},
Linlin Xu\textsuperscript{2}
David Clausi\textsuperscript{1},
\\
\\
\textsuperscript{1}Systems Design Engineering, University of Waterloo \\
\textsuperscript{2}Department of Geomatics Engineering,University of Calgary \\
{\tt\small (m32patel, jnoaturnes, j3hsiao, dclausi)@uwaterloo.ca,
lincoln.xu@ucalgary.ca}
}

\begin{document}
\maketitle

\begin{abstract}
We introduce OpenWildlife (OW), an open-vocabulary wildlife detector designed for multi-species identification in diverse aerial imagery. While existing automated methods perform well in specific settings, they often struggle to generalize across different species and environments due to limited taxonomic coverage and rigid model architectures. In contrast, OW leverages language-aware embeddings and a novel adaptation of the Grounding-DINO framework, enabling it to identify species specified through natural language inputs across both terrestrial and marine environments. Trained on 15 datasets, OW outperforms most existing methods, achieving up to \textbf{0.981} mAP50 with fine-tuning and \textbf{0.597} mAP50 on seven datasets featuring novel species. Additionally, we introduce an efficient search algorithm that combines k-nearest neighbors and breadth-first search to prioritize areas where social species are likely to be found. This approach captures over \textbf{95\%} of species while exploring only \textbf{33\%} of the available images. To support reproducibility, we publicly release our source code and dataset splits, establishing OW as a flexible, cost-effective solution for global biodiversity assessments.
\end{abstract}
    
\begin{figure*}[!ht]
    \centering
    \includegraphics[width=\linewidth]{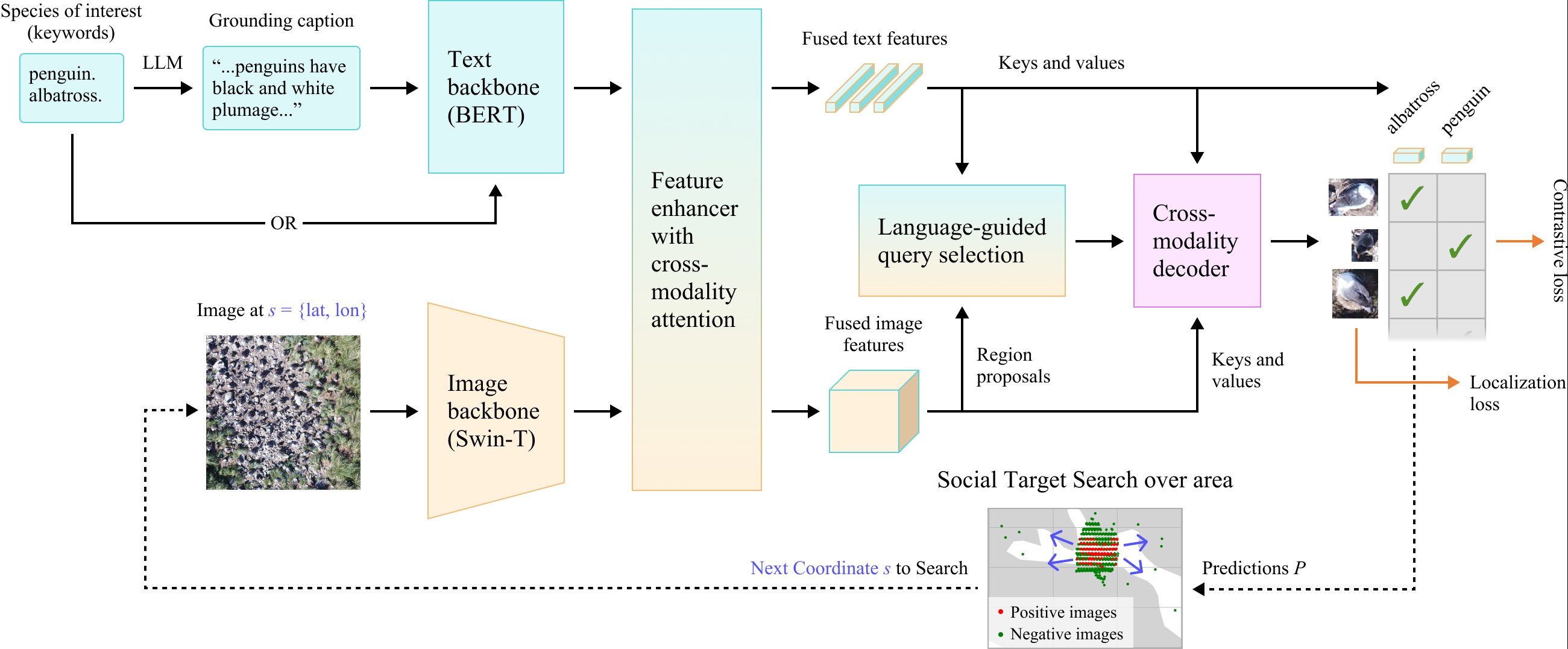}
    \caption{In the OpenWildlife architecture, given an aerial image and either a list of species of interest or an LLM-generated grounding caption, the model extracts image and text features. These features are fused through cross-modality enhancement, enabling language-guided queries to highlight relevant regions in both modalities. Predicted animal locations can then be fed into the Social Target Search (STS) module, which prioritizes the next regions for detection based on potential social aggregations of target species. This iterative process efficiently pinpoints wildlife across large areas spanning multiple images.}
    \label{fig:architecture_and_process}
\end{figure*}

\section{Introduction}
\label{sec:intro}

Monitoring wildlife is essential for understanding ecosystem health and guiding conservation efforts \cite{linchantAreUnmannedAircraft2015}. To monitor species populations, remote sensing has proven to be an effective approach \cite{Skidmore2021PrioritySpace, Leyequien2007CapturingDiversity}. Aerial imagery captured with aircrafts can be retrospectively annotated to count the population of species present \cite{Delplanque2024WildlifeYet, Cubaynes2022WhalesModels}. This annotation is commonly conducted manually, an approach that is both costly and prone to subjectivity \cite{Lamprey2020CamerasUganda, Bowler2020UsingUncertainty}. Automated methods using artificial intelligence have been able to surpass manual counting in speed and reliability \cite{Torney2019AImages, Hodgson2016PrecisionVehicles, Norouzzadeh2018AutomaticallyLearning, Kellenberger2018DetectingLearning}. Recent models such as HerdNet \cite{Delplanque2023FromLearning} and WildlifeMapper \cite{Kumar2024WildlifeMapperIdentification} have excelled in detecting multiple species despite heavy occlusion on complex, heterogeneous backgrounds. However, these models are limited in their ability to generalize to novel species or known species in novel environments — an important ability for effective conservation \cite{reynoldsPotentialAIRevolutionize2025}. 

Additionally, while advancements in AI models have improved detection capabilities, there are still challenges in the collection of aerial images themselves. Surveys for aerial imagery face high operational costs, limited coverage, and various logistical constraints such as airspace regulations and weather conditions. With the recent advancements in satellite technology, space-based imagery has become an increasingly viable alternative for wildlife monitoring. Future conservation efforts could rely on on-demand satellite imagery, offering a cost-effective solution to supplement or even replace traditional field campaigns \cite{Delplanque2024WildlifeYet}. In such scenarios, it becomes crucial to efficiently select regions likely to be inhabited by target species, minimizing the number of satellite images required for monitoring.

This paper introduces OpenWildlife (OW), an open-vocabulary wildlife detector for geographically-diverse aerial imagery, and Social Target Search (STS), an algorithm to propose locations where wildlife is likely to be found. Combined, these methods will enable conservation agencies to detect novel species without the need to manually annotate training data and optimize image acquisition by strategically selecting regions for analysis. We also release a large-scale annotated whale dataset collected by Fisheries and Oceans Canada (DFO) using camera-equipped aircraft over North-American Arctic waters from 2014 to 2017. In summary, our contributions are as follows:

\begin{enumerate}
    \item A novel, end-to-end approach for open-vocabulary multi-species animal detection from aerial imagery. By categorizing species in a language-aware semantic space, our approach improves upon existing methods in its ability to generalize to novel species as well as known species in novel environments (\cref{sec:openwildlife_architecture}).
    
    \item A novel search method that combines K-Nearest Neighbors (KNN) and Breadth-First Search (BFS) to propose geographic locations likely to contain animals, leveraging species' social behavior. Tested on geo-referenced datasets, this method efficiently captures a significant proportion of wildlife-populated images while exploring only small portions of the region (\cref{sec:search_algorithm}).
    
    \item A comprehensive benchmark based on publicly available aerial wildlife imagery datasets (\cref{sec:consolidated_dataset}), including the large-scale whale monitoring dataset from DFO (\cref{sec:dfowhale_dataset}).
\end{enumerate}
\section{Related Works}
\label{sec:related_works}

\noindent \textbf{Manual Animal Detection:}  
The standard method for wildlife counting over large areas is the systematic reconnaissance flight (SRF), where rear-seat-observers (RSOs) count animals in sample strips \cite{Lamprey2020ComparingLevel, Eltringham1977CountingAnimals.}. The primary limitation is the brief timeframe for counting, often leading to underestimation \cite{Caughley1974BiasSurvey}. While replacing real-time observation with retrospective enumeration can improve accuracy, manual counting remains prone to error \cite{Leedy1948AerialManagement, Lamprey2020CamerasUganda, Bowler2020UsingUncertainty}.

\noindent \textbf{Automated Animal Detection:}  
Animal detection from aerial imagery is challenging due to occlusion, non-uniform distribution, camouflage, and varying scale \cite{Delplanque2023FromLearning, Eikelboom2019ImprovingDetection, Patel2023TheImagery, Jachmann2002ComparisonHerbivores}. Modified U-Net architectures \cite{Ronneberger2015U-Net:Segmentation} have enabled high-precision detection of terrestrial mammals from satellite and drone imagery \cite{Wu2023DeepLandscape, Han2019LivestockNetwork, Goncalves2020SealNet:Imagery}. To improve precision in animal censuses, where most images lack animals, techniques like under-sampling, hard negative mining, and introducing a border class have been proposed \cite{Patel2023TheImagery, Kellenberger2018DetectingLearning}. \cite{Delplanque2023FromLearning} used a density-based encoder-decoder for locating and enumerating herds, while \cite{Kumar2024WildlifeMapperIdentification} augmented a ViT-based encoder with a high-frequency feature generator to highlight objects of interest against complex backgrounds.

\noindent \textbf{Open-set Object Detection:}  
The problem of open-set recognition, where models encounter unknown classes at test time, was first proposed by \cite{Scheirer2013TowardRecognition}. \cite{Dhamija2020TheSet} formalized open-set object detection (OSOD), where models only classify known objects. However, OSOD models require additional human annotation for novel categories. \cite{Zheng2022TowardsDiscovery} extended OSOD to discover novel categories based on visual similarity to known ones. This has expanded to open-vocabulary object detection, where novel objects are described by arbitrary natural language inputs. Approaches like \cite{Zang2022Open-VocabularyMatching, Gu2021Open-vocabularyDistillation} leverage CLIP embeddings for region-guided learning, and \cite{Li2021GroundedPre-training} uses phrase grounding to align semantic understanding between text and regions.

\noindent \textbf{Target Search Strategies:} To the best of our knowledge, there is limited literature in remote sensing focused on searching for social targets. While approaches related to tracking dynamic targets using multi-agent systems and optimization methods for search and rescue \cite{Yan2021, Cao2024} are somewhat relevant, our approach of searching for targets based on their social behavior is novel (see \cref{sec:search_algorithm}).

\section{OpenWildlife Model}
\label{sec:openwildlife_architecture}

\Cref{fig:architecture_and_process} illustrates the architecture of our OpenWildlife model, which builds upon MM-Grounding-DINO \cite{Zhao2024AnDetection} to enable aerial wildlife detection by integrating vision and language features. The model consists of five key components: caption generation, feature extraction and fusion, language-guided query selection, cross-modality decoding, and training losses, which are detailed below.

\subsection{Caption Generation}
A key challenge in applying Grounding-DINO to aerial wildlife detection is the lack of semantically rich image-caption datasets for this domain. While Grounding-DINO is trained on datasets like O365 \cite{Shao2019Objects365:Detection} and V3Det \cite{Wang2023V3Det:Dataset}, no equivalent dataset exists for aerial wildlife. To address this, we optionally use large language models (LLMs) to generate captions tailored to aerial wildlife imagery. These captions provide species descriptions that guide detection, serving as inputs to the text encoder. The caption generation process is detailed further in \cref{sec:caption_generation}.

\subsection{Feature Extraction and Fusion}
Once captions are generated, they are processed by a BERT-based text encoder\cite{devlin2019bert}, while a Swin Transformer\cite{liu2021swin} backbone extracts multi-scale visual features from high-resolution aerial images. The image backbone captures both fine-grained object details and broader contextual patterns. These representations are then fused in a feature enhancer module, which employs bi-directional attention to align text and image features, followed by self-attention and deformable attention layers to refine representations.

\subsection{Language-Guided Query Selection}
To efficiently localize wildlife, the language-guided query selection module generates region proposals by computing cosine similarity between visual features and species-specific text embeddings. This mechanism ensures that the model prioritizes regions most relevant to the species described in the caption. The selected queries are then used as input to the decoder for further refinement.

\subsection{Cross-Modality Decoder}
The cross-modality decoder refines object localization by iteratively incorporating visual and textual information. Each decoder layer consists of self-attention, image cross-attention, and text cross-attention layers, followed by a feedforward network (FFN). Compared to the standard DINO decoder, MM-Grounding-DINO includes an additional text cross-attention layer per decoder block to inject more linguistic context into the object queries. This ensures that species names and descriptions effectively influence the bounding box predictions.

\subsection{Training Losses}
We adopt multiple loss functions to optimize the model. For bounding box regression, we use L1 loss and GIoU loss \cite{rezatofighiGeneralizedIntersectionUnion2019}. For classification, we employ focal loss \cite{linFocalLossDense2018} as a contrastive loss between predicted boxes and language tokens, ensuring that detected objects align with text descriptions. This contrastive loss is the key to achieving zero-shot detection on novel species\cite{radford2021learning}. Additionally, we incorporate bipartite matching loss \cite{carionEndtoEndObjectDetection2020} to improve correspondence between ground truth and predicted boxes. As in DETR \cite{carionEndtoEndObjectDetection2020}, we also apply auxiliary loss at each decoder layer.

\section{Social Target Search}
\label{sec:search_algorithm}

A large variety of species naturally form complex social structures, interact with each other regularly, and often live in groups \cite{Ward2016Sociality:Animals}. This prior knowledge is important for wildlife monitoring, as it allows the identification of geographic locations that are more likely to contain target species. Leveraging this behavior, we designed the Social Target Search (STS) algorithm (\Cref{algor:sts}) to identify regions within a larger survey area where target species are likely to be found. These identified regions can then be prioritized for further manual or automated investigation, effectively minimizing the amount of imagery to locate target species populations. Overall, STS begins by randomly sampling images until target animals are detected, then gradually expands to nearby images assuming higher likelihood of containing wildlife.

\begin{algorithm}[!htb]
\caption{STS: Social Target Search}
\label{algor:sts}
\begin{algorithmic}[1]

\State \textbf{Input:} 
     \State $OW$     \Comment{Detection model}
     \State $C = \{(lat, long)\}$     \Comment{Coordinates of image centers}
     \State $ k \in \mathbb{Z}^+ \quad | \quad k \ll |C|$         \Comment{Number of Samples}
     \State $d \in \mathbb{R}^+$      \Comment{Maximum neighbor distance}
\State \textbf{Output:} 
     \State $n = 0$ \Comment{Number of images evaluated}

\Algphase{Phase 1 - Sampling images without replacement}

\State $Q \gets \emptyset$ empty queue
\Repeat
    \State $S \leftarrow \text{Sample}(C, k) \quad \text{where} \quad S \subseteq C, \quad |S| = k$
    \State $n \gets n + |S|$
    \For{\textbf{each } $s \in S$}
        \State $I \gets $ Acquire image centered at $s$
        \State $P \gets model(I)$   \Comment{Predictions}
        \If{$P \neq \emptyset$}
            \State $Q \gets Q \cup \{ s \}$
        \EndIf
    \EndFor
    \State $C \gets C \setminus S$  
\Until{$Q \neq \emptyset$}

\Algphase{Phase 2 - BFS with nearest neighbor graph strategy}

\While{$Q \neq \emptyset$}

    \State $KDTree \gets $ Create tree from coordinates in $C$
    \State $q \gets Q.dequeue()$    \Comment{Parent node}
    \State $S \gets KDTree.query(q, k, d)$   \Comment{Neighbors}

    \State $n \gets n + |S|$
    \For{\textbf{each } $s \in S$}
        \State $I \gets $ Acquire image centered at $s$
        \State $P \gets OW(I)$   \Comment{Predictions}
        \If{$P \neq \emptyset$}
            \State $Q \gets Q \cup \{ s \}$
        \EndIf
    \EndFor
    \State $C \gets C \setminus S$ 

\EndWhile
\AlgphaseAux{}
\State return $n$

\end{algorithmic}
\end{algorithm}

\subsection{Algorithm Description}

Initially, we define a set $C$ containing all possible latitude and longitude pairs $(lat, long)$ as the potential image centers within our search area. We also specify the number of sample images $k$ to be evaluated in each iteration, as well as the maximum distance $d$ within which two images are considered neighbors. The searching consists of two stages. First, we randomly sample $k$ coordinate pairs from the set of locations $C$. For each sampled coordinates $s$ we obtain the corresponding remote sensing image and predict the location of the target wildlife. If at least one animal is detected (predictions $P \neq \emptyset$), the correspondent pair $s$ is pushed into a queue $Q$ that will contain the location coordinates from populated images. All sampled coordinates are removed from the initial set $C$ to ensure future sampling without replacement. This process is repeated until $Q$ contains at least one coordinate pair.

Starting with at least one location from a populated image, the sampling process is guided by a BFS method, which explores neighboring locations in $C$ relative to coordinates dequeued from $Q$. For each newly sampled location, coordinates whose images contain a target animal are pushed into $Q$, and the process is repeated until $Q$ is empty. All visited coordinates are continuously excluded from $C$. Since neighboring coordinates can be located in any direction or at any distance, the KNN method is used as a criterion to find adjacent nodes. To enhance efficiency, a space-partitioning binary tree, known as a KD-Tree, is constructed during each BFS iteration using the current coordinates in $C$. The construction complexity of the KD-Tree is $O(|C|\log|C|)$, but it becomes less computationally demanding as the cardinality of $C$ decreases. Neighbors are retrieved by querying the KD-Tree, with a complexity of $O(\log|C|)$, to find the $k$ nearest locations to the parent coordinates $q$ that are within a specified distance $d$.

\subsection{Limitations}

The STS may converge to a local area if the initial set of coordinates associated with populated images is located farther than the distance $d$ from neighboring populated images, which could happen when a few animals are separated from the herd. In that scenario, repeating phases 1 and 2 is sufficient to increase the possibility to find new groups.

\section{DFO Whale Dataset}
\label{sec:dfowhale_dataset}

The DFO whale dataset comprises aerial surveys conducted over four years (2014 \cite{MarcouxEstimateSurvey}, 2015 \cite{Matthews2017CanadianSurvey}, 2016 \cite{Marcoux2019CanadianSurvey}, and 2017 \cite{Watt2021AbundancePopulation}) by Fisheries and Oceans Canada (DFO) to monitor critical habitats for beluga (\textit{Delphinapterus leucas}) and narwhal (\textit{Monodon monoceros}) populations. The 2014 and 2017 surveys focused on high-density beluga aggregations in Cumberland Sound, the 2015 survey covered Western Hudson Bay, and the 2016 survey targeted narwhal distribution in Eclipse Sound. Surveys were conducted at altitudes of $1,000$ to $2,000$ feet using a DeHavilland Twin Otter aircraft equipped with a $36.15$ MP Nikon D810 camera and GPS for geo-referencing.

\begin{figure}[!htb]
    \centering
    \includegraphics[width=\linewidth]{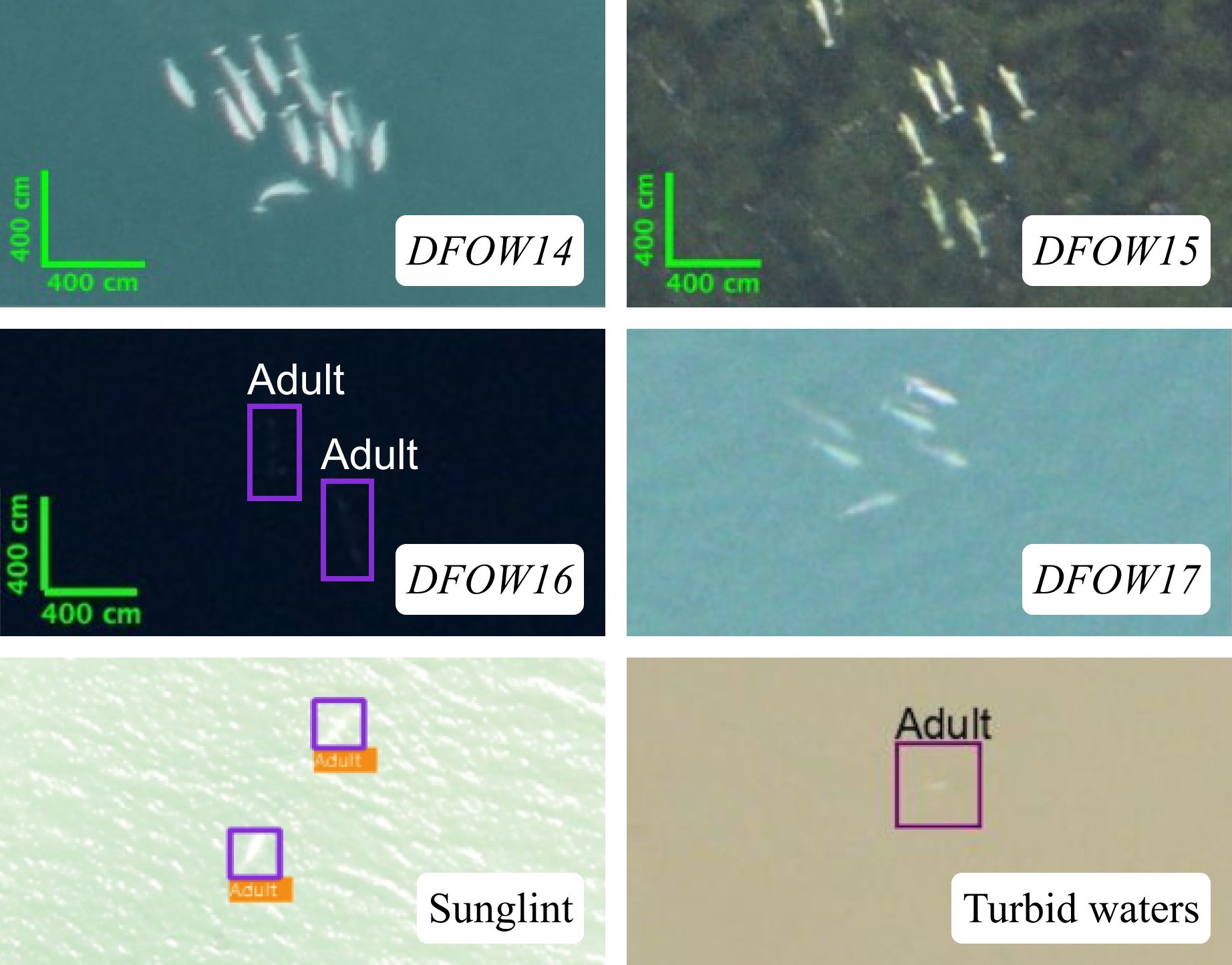}
    \caption{Examples of the DFO whale dataset, highlighting environmental variations across survey years and the impact of sunglint and turbid water on beluga detectability in aerial imagery.}
    \label{fig:DFO dataset}
\end{figure}

This dataset includes $1,017$ GB of data, with $39,492$ images and $58,292$ whale detections in $1,644$ images. Each image is annotated by DFO experts with categories: Adult, Calf, Juvenile, Unknown, and Questionable. The images have a resolution of $4912\times 7360$ pixels, with an average spatial resolution of $10$ cm/pixel. Annotations are converted to $40 \times 40$ pixel bounding boxes ($400 \times 400$ cm), based on the average size of belugas and narwhals ($4$ meters) \cite{Ocorry-Crowe2009BelugaLeucas}. Table \ref{tab:summary_whale_findings} summarizes the whale annotations. Visibility is affected by factors like time of day, wind speed, viewing angle, sun glare, and water turbidity, as illustrated in \Cref{fig:DFO dataset}.

\begin{table}[!htb]
    \centering
    \resizebox{\columnwidth}{!}{
    \begin{tabular}{l l | c  >{\centering\arraybackslash}p{1.7cm}  c}
        \textbf{Dataset} & \textbf{Species} & \textbf{Images} & \textbf{Images w/ Whales} & \textbf{Whales} \\
        \rowcolor{gray!10}
        \hline
        DFOW14 & Beluga & 18,718 & 467 & 1,935 \\
        DFOW15 & Beluga & 5,440 & 278 & 47,459 \\
        \rowcolor{gray!10}
        DFOW16 & Narwhal & 10,102 & 666 & 5,181 \\
        DFOW17 & Beluga & 5,232 & 233 & 3,717 \\
        \hline
        \textbf{Total} & & \textbf{39,492} & \textbf{1,644} & \textbf{58,292} \\
    \end{tabular}
    }
    \caption{Summary of images and annotations from different DFO surveys.}
    \label{tab:summary_whale_findings}
\end{table}

\begin{table*}
    \centering
    \resizebox{\linewidth}{!}{
        \begin{tabular}{p{3cm}<{\raggedright} | p{8cm}<{\raggedright} l l}
             & Species labels & \# of annotated images & \# of annotations\\
             \hline
             \rowcolor{gray!10}
             Train & Alcelaphinae, beluga, bird, brant, buffalo, Canada goose, elephant, emperor goose, kob, penguin, sea lion, turtle, warthog, waterbuck & 27,684 & 977,428\\
             Test (novel species) & Camelus, cattle, gray headed gull, great cormorant, great white pelican, gull, kiang, narwhal, polar bear, royal tern, seal, sheep, slender-billed gull, yak, zebra & 22,028 & 320,381\\
             \rowcolor{gray!10}
             Test (novel environments) & Beluga whale, bowhead whale, elephant, giraffe, narwhal, penguin, zebra & 1,226 & 8,360\\
        \end{tabular}
    }
    \caption{Dataset partitioning for open-set continuous pretraining and evaluation. The test (novel species) set contains 15 species that do not appear in the training set, ensuring no overlap in species for open-set evaluation. The test (novel environments) set, in contrast, includes seven species already present in the dataset but captured in new environments, enabling evaluation on domain generalization.}
    \label{tab:dataset_split}
\end{table*}

\section{Experimental Setup}
\label{sec:experimental_setup}

\subsection{Consolidated Dataset}
\label{sec:consolidated_dataset}

We consolidated 25 publicly available aerial wildlife datasets for our study. Each dataset was converted into the COCO annotation format with bounding boxes; datasets containing only key points were adapted by generating fixed-size bounding boxes. The datasets were divided into three groups: 15 for open-set continuous pretraining, 7 for evaluating novel species detection, and 3 for assessing model performance in novel environments. Details of the species in each group are provided in \cref{tab:dataset_split} of the supplementary material. This partitioning is designed to evaluate the model's ability to detect both unseen species and familiar species in unfamiliar environments — a common challenge in aerial surveys. Additionally, when determining the splits, we ensured a fair representation of wildlife types (e.g., bird, marine, terrestrial) across all groups.

\subsection {Caption Generation}
\label{sec:caption_generation}

We trained two versions of the OpenWildlife model: one using LLM-generated captions ("OW Sentence") and the other using a keyword list of species present in the image ("OW Keyword").

For OW Sentence, we use OpenAI’s GPT-4o-mini \cite{OpenAI2024GPT-4oOpenAI} Batch API to generate a unique caption for each image, ensuring alignment with the provided bounding box labels. Each caption begins with a species list matching the labeled instances, followed by a brief visual description based on the fine-grained species in the source dataset. The descriptions focus on broad, aircraft-visible features such as color and body shape (e.g., "zebras with black and white stripes"). The prompt, customized per image with its corresponding labels and true species, is as follows: \textit{"You are an annotator generating captions of an aerial image. The caption should start with the species present, following exactly the LABELS, followed by a very short description of their appearance based on the TRUE SPECIES. The TRUE SPECIES should not be mentioned, only their descriptions. If LABELS is 'NO INSTANCES,' only say 'No animals present.' Only describe details visible from very far away."}. Examples of generated captions are available in Supp. \cref{tab:dataset_captions}.

\subsection{Implementation and Training}

Our OpenWildlife (OW) model was initialized with publicly available MM-GDINO-T weights \cite{Zhao2024AnDetection}. Since most datasets consist of high-resolution aerial images, training images were pre-sliced into $1024 \times 1024$ pixel patches. During testing, images were sliced in real time and processed individually, with results merged via non-maximum suppression (NMS). All reported metrics are based on full-scale test images rather than individual slices.

The open-set continuous pretraining of OW was conducted on the 15 datasets detailed in Supp. \cref{tab:training_datasets}. To ensure fair comparison with existing work, we used the original train/test/val splits when explicitly provided in the literature. For datasets without predefined splits (denoted with $\ddag$), we replicated the methodology described in the corresponding studies to create our own, ensuring consistency across evaluations. 

A single OW model was jointly trained on all train datasets for 20 epochs and subsequently evaluated on the test portion of each dataset. Training was performed on an HPC cluster with 8 NVIDIA A100-40GB GPUs across 2 nodes, using a global batch size of $16$. The full 20-epoch training cycle took approximately $20$ hours.

Since evaluation protocols vary across studies, we report results using both the metrics commonly found in the literature and the standard mean Average Precision (mAP) at an intersection-over-union (IoU) threshold of 0.50.
\section{Results}
\label{sec:results}

\subsection{Literature Comparison}

As shown in \Cref{tab:train_results}, OW outperforms most existing methods, with the exception of the Qian and Virunga-Garamba datasets. The Qian dataset, featuring densely packed penguin colonies, is better suited for crowd-counting regression approaches like \cite{Qian2023CountingModel}, which handle occlusion more effectively \cite{liu2018decidenet}. For Virunga-Garamba, OW was slightly outperformed by \cite{Kumar2024WildlifeMapperIdentification}, likely due to their use of the SAM ViT-L backbone \cite{kirillov2023segment} with 308M parameters, compared to OW’s lighter model with 172M parameters.




\begin{table}[!htb]
    \centering
    \resizebox{\columnwidth}{!}{
    \begin{tabular}{@{} l | p{2.75cm}<{\raggedright} | p{2cm}<{\raggedright} l @{}}
        \textbf{Dataset} & \textbf{Literature Results} & \multicolumn{2}{c}{\textbf{OW Keyword}} \\ 
        & & Lit. Metric & mAP50 \\
        \hline
        \rowcolor{gray!10}
        IndOcean \cite{Weinstein2022AImagery}& $0.72$ F1 \cite{Weinstein2022AImagery} & $\mathbf{0.80}$ F1 & $0.833$ \\ 
        NewMex \cite{Weinstein2022AImagery} & $0.76$ F1 \cite{Weinstein2022AImagery} & $\mathbf{0.826}$ F1 & 0.728 \\ 
        \rowcolor{gray!10}
        Palmyra \cite{Weinstein2022AImagery} & $0.72$ F1 \cite{Weinstein2022AImagery} & $\mathbf{0.78}$ F1 & 0.789 \\ 
        Pfeifer \cite{pfeifer2021assessing} & $0.66$ F1 \cite{Weinstein2022AImagery} & $\mathbf{0.86}$ F1 & $0.842$ \\ 
        \rowcolor{gray!10}
        Seabird & $0.72$ F1 \cite{Weinstein2022AImagery} & $\mathbf{0.93}$ F1 & $0.959$ \\ 
        Qian \cite{Qian2023CountingModel} & $\mathbf{66.27}$ RMSE \textsuperscript{\ddag} & $85.32$ RMSE & $0.671$ \\
        \rowcolor{gray!10}
        WAID \cite{Mou2023WAID:Drones} & $\mathbf{0.98}$ mAP50 \cite{Mou2023WAID:Drones} & $\mathbf{0.98}$ mAP50 & $0.981$ \\ 
        AED \cite{Naude2019TheDetection.} & $0.89$ mAP \textsuperscript{\dag} \cite{Naude2019TheDetection.} & $\mathbf{0.90}$ mAP \textsuperscript{\dag} & $0.856$ \\ 
        \rowcolor{gray!10}
        Virunga-Gar.\cite{Delplanque2022MultispeciesNetworks} & $\mathbf{0.85}$ mAP50 \cite{Kumar2024WildlifeMapperIdentification}, 0.835 F1 \cite{Delplanque2023FromLearning}& $0.72$ mAP50, $\mathbf{0.85}$ F1 & $0.719$ \\ 
        Sea Lion \cite{Kaggle2017NOAAKaggle} & $10.86$ RMSE \cite{Kaggle2017NOAAKaggle} & $\mathbf{8.13}$ RMSE & $0.662$ \\ 
        \rowcolor{gray!10}
        Turtle \cite{Gray2019AImagery} & $0.27$ F1 \textsuperscript{\ddag} & $\mathbf{0.38}$ F1 & $0.26$ \\ 
        Izembek \cite{Weiser2023OptimizingCounting} & - & - & $0.343$ \\ 
        \rowcolor{gray!10}
        DFOW14 & - & - & $0.856$ \\ 
        DFOW15 & - & - & $0.82$ \\ 
        \rowcolor{gray!10}
        DFOW17 & - & - & $0.857$ \\ 
    \end{tabular}
    }
    \caption{Results of open-set continuous fine-tuning. \textbf{Bold} numbers denote the best metric between literature results and OW Keyword results. Results marked with $\dagger$ indicate a 200-pixel Chebyshev distance allowance to be considered a true positive. Literature results marked with $\ddagger$ were reproduced by us due to the original splits not being available.}
    \label{tab:train_results}
\end{table}

\subsection{Detection of Novel Species}

\begin{figure}[h]
    \centering
    \includegraphics[width=\linewidth]{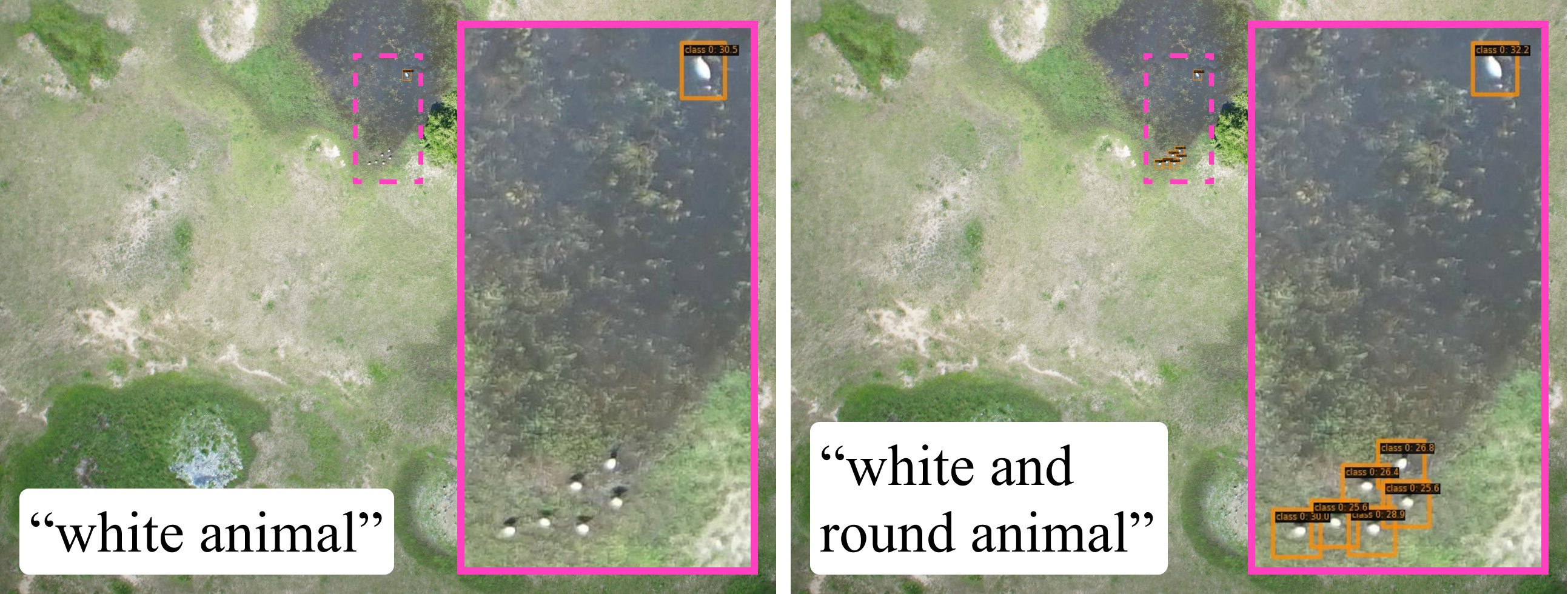}
    \caption{Qualitative detection results on Virunga-Garamba for different text prompts. OW exhibits semantic understanding of shape and color.}
    \label{fig:semantic_understanding}
\end{figure}

We evaluate detection of novel species on seven unseen datasets, with the results presented in \Cref{tab:novel_species_results}. As seen in the table, OW exhibits fair novel species detection performance (up to 0.597 mAP50) and outperforms most literature results when fine-tuned only for 10 epochs. OW Sentence outperforms OW Keyword when the sentence caption is sufficiently descriptive, improving bounding box precision and robustness to background features. This is due to OW’s ability to leverage semantic understanding, as demonstrated in \Cref{fig:semantic_understanding}. Descriptive captions that highlight shape and other contextual elements enable OW to better align with image features and reduce class confusion. However, when the LLM-generated caption is too long or inaccurate, this can hinder the model’s performance, as seen in some failure cases. Qualitative insights into OW's multi-species detection capability in complex terrestrial and marine scenarios are shown in \Cref{fig:full_zero_shot_results}. Additional qualitative results are presented in supplemental materials.

\begin{table}[!htb]
    \centering
    \resizebox{\columnwidth}{!}{
    \begin{tabular}{@{} l | p{1.4cm}<{\raggedright} p{1.4cm}<{\raggedright} | p{1.15cm}<{\raggedright} p{1.15cm}<{\raggedright} p{1.4cm}<{\raggedright} @{}}
        \multirow{2}{*}{\textbf{Dataset}} & \multicolumn{2}{c|}{\textbf{Novel Species}} & \multicolumn{3}{c}{\textbf{Finetune}} \\ 
        & \textbf{Sentence (mAP50)} & \textbf{Keyword (mAP50)} & \textbf{Lit. Results} & \textbf{Lit. Metric} & \textbf{Keyword (mAP50)} \\ 
        \hline
        \rowcolor{gray!10}
        WAfrica & $0.465$ & $\mathbf{0.533}$ & $0.79$ F1 \cite{Kellenberger202121000Learning} & $\mathbf{0.91}$ F1 & $0.908$ \\ 
        Michigan & $\mathbf{0.179}$ & $0.131$ & $0.72$ F1 \cite{Weinstein2022AImagery} & $\mathbf{0.82}$ F1 & $0.813$ \\ 
        \rowcolor{gray!10}
        DFOW16 & $0.576$ & $\mathbf{0.597}$ & - & - & $0.805$ \\ 
        Polar Bear & $\mathbf{0.369}$ & $0.224$ & - & - & $0.716$ \\ 
        \rowcolor{gray!10}
        ArctSeal & $\mathbf{0.257}$ & $0.183$ & $\mathbf{0.87}$ F1 \textsuperscript{\ddag} & $\mathbf{0.87}$ F1 & $0.916$ \\ 
        Han & $0.106$ & $\mathbf{0.161}$ & $0.89$ mAP50 \cite{Han2019LivestockNetwork} & $\mathbf{0.96}$ mAP50 & $0.964$ \\ 
        \rowcolor{gray!10}
        WAID & $\mathbf{0.229}$ & $0.212$ & $\mathbf{0.98}$ mAP50 \cite{Mou2023WAID:Drones} & $0.97$ mAP50 & $0.967$ \\ 
    \end{tabular}
    }
    \caption{Results on novel species datasets. \textbf{Bold} numbers highlight the best metric, either between the OW Keyword and OW Sentence models, or comparing literature results with OW Keyword results. Literature results marked with $\ddagger$ were reproduced by us due to the original splits not being available.}
    \label{tab:novel_species_results}
\end{table}

\subsection{Detection in Novel Environments}

We evaluate detection in novel environments on a test set comprising seven species already present in the training dataset but captured in new environments, enabling assessment of domain generalization. As seen in the results presented in \Cref{tab:novel_domain_results}, OW achieves up to 0.733 mAP50 for domain transfer and outperforms literature results when fine-tuned only for 10 epochs.

\begin{table}[!htb]
    \centering
    \resizebox{\columnwidth}{!}{
    \begin{tabular}{@{} l | p{1.4cm}<{\raggedright} p{1.4cm}<{\raggedright} | p{1.15cm}<{\raggedright} p{1.15cm}<{\raggedright} p{1.4cm}<{\raggedright} @{}}
        \multirow{2}{*}{\textbf{Dataset}} & \multicolumn{2}{c|}{\textbf{Novel Environments}} & \multicolumn{3}{c}{\textbf{Finetune}} \\ 
        & \textbf{Sentence (mAP50)} & \textbf{Keyword (mAP50)} & \textbf{Lit. Results} & \textbf{Lit. Metric} & \textbf{Keyword (mAP50)} \\ 
        \hline
        \rowcolor{gray!10}
        Penguin & $\mathbf{0.733}$ & $0.619$ & - & - & $0.755$ \\ 
        DFOW23 & $0.024$ & $\mathbf{0.362}$ & - & - & $0.435$ \\ 
        \rowcolor{gray!10}
        Kenya & $\mathbf{0.384}$ & $0.326$ & $0.77$ mAP30 \cite{Eikelboom2019ImprovingDetection} & $\mathbf{0.869}$ mAP30 & $0.854$ \\ 
    \end{tabular}
    }
    \caption{Results on novel environment datasets. \textbf{Bold} numbers highlight the best metric, either between the OW Keyword and OW Sentence models, or comparing literature results with OW Keyword results.}
    \label{tab:novel_domain_results}
\end{table}

\begin{figure*}[h]
    \centering
    \includegraphics[width=\linewidth]{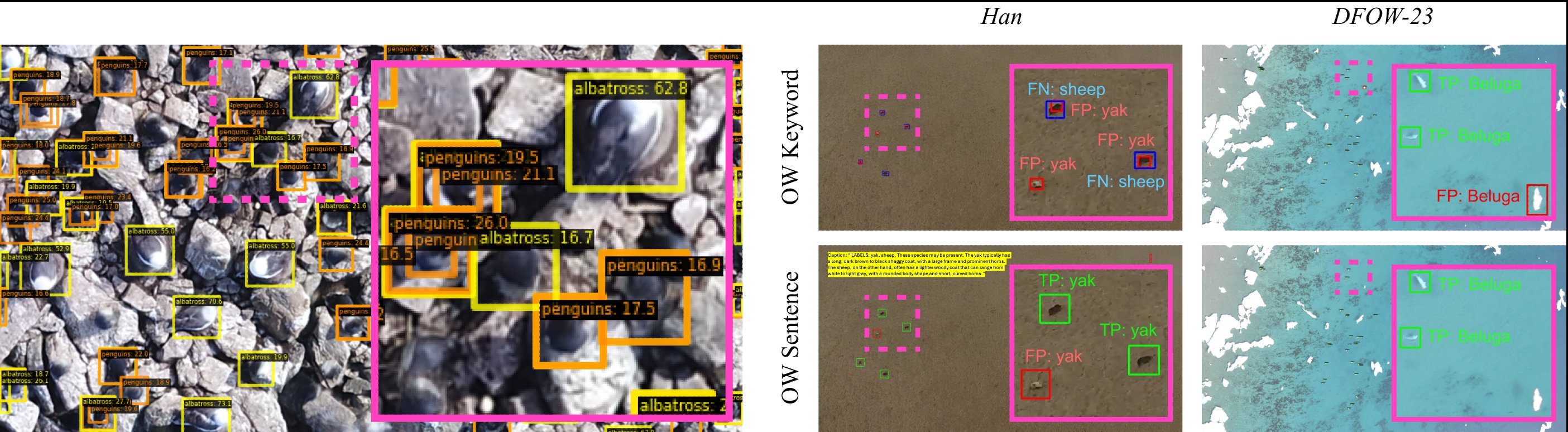}
    \caption{Qualitative detection results on novel species and in novel environments. Left: multi-species detection. Right: contrasts the performance of the OW Keyword and OW Sentence models. TP is marked as green, FP as red, FN as blue. Grounding with descriptive sentences can reduce class confusion and improve robustness to background features.}
    \label{fig:full_zero_shot_results}
\end{figure*}

\begin{figure}[h]
    \centering
    \includegraphics[width=\linewidth]{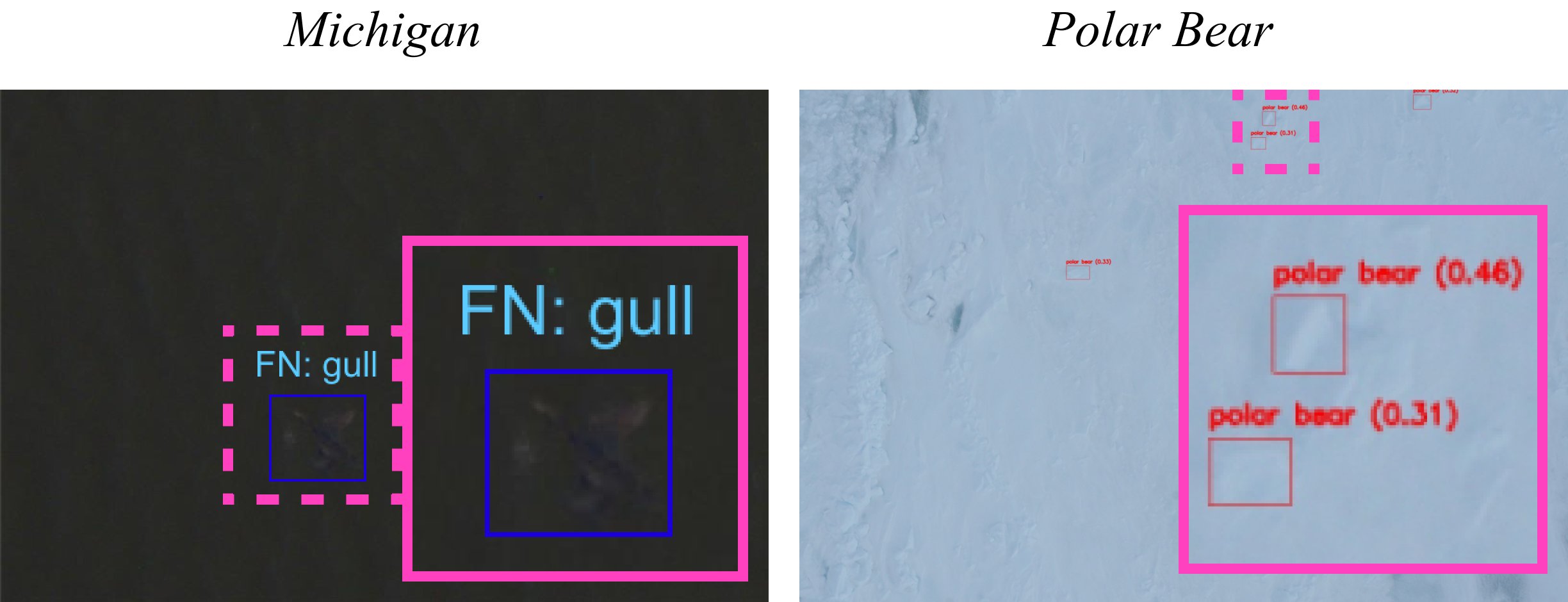}
    \caption{Limitations of novel species detection. Left: OW Sentence failed to detect a gull because the LLM-generated caption described it as "white", not "brown". Right: OW classified a snow mound as a polar bear.}
    \label{fig:ow_limitations}
\end{figure}

\subsection{Limitations}

In novel species detection, OW may struggle when LLM-generated captions provide misleading descriptions. For example, OW failed to detect a gull because the caption described it as "white" instead of "brown". However, after correcting the description, OW was able to correctly detect the gull. Additionally, OW misclassified a snow mound as a polar bear, likely because it had not encountered that background element during training. Visualized failure cases are shown in \cref{fig:ow_limitations}.

\subsection{STS Evaluation}

To evaluate the Social Target Search (\cref{algor:sts}), we used location data from all images in the different surveys of the DFO whale dataset. Due to the stochastic nature of phase 1, average metrics on $20$ runs were reported for each experiment. The number of samples/neighbors was set up at $k=10$. The maximum distance for KNN was set to $d=0.6^\circ$, which represents approximately $66$ km in latitude and between $19$ km and $27$ km in longitude. We use random search as baseline. To achieve a fair comparison, both STS and random search are run until $95\%$ of the images containing whales are found.

\begin{table}[!htb]
    \begin{tabular}{c | l | c | c }
        \multirow{2}{*}{\textbf{Dataset}}    &   \multirow{2}{*}{\textbf{Method}}  &    \multirow{2}{1.8cm}{\centering \textbf{\% Images Analyzed} $\downarrow$}   &   \multirow{2}{1.8cm}{\centering\textbf{\% Whales Detected} $\uparrow$} \\
        &&&\\
        \hline
        \rowcolor{gray!10}
        \multirow{2}{*}{\cellcolor{white} \raggedright DFOW14}
        &   STS     &   $13.8\pm1.3$   &   $98.5\pm0.3$   \\
        &   Rand S  &   $94.5\pm1.5$   &   $95.5\pm2.3$   \\

                \hline
        \rowcolor{gray!10}
        \multirow{2}{*}{\cellcolor{white} \raggedright DFOW15}
        &   STS     &   $36.1\pm4.6$   &   $99.4\pm0.6$   \\
        &   Rand S  &   $95.4\pm1.2$   &   $94.9\pm2.3$   \\

        \hline
        \rowcolor{gray!10}
        \multirow{2}{*}{\cellcolor{white} \raggedright DFOW16}
        &   STS     &   $47.4\pm1.4$   &   $95.2\pm5.1$   \\
        &   Rand S  &   $94.8\pm0.9$   &   $95.1\pm1.4$   \\

        \hline
        \rowcolor{gray!10}
        \multirow{2}{*}{\cellcolor{white} \raggedright DFOW17}
        &   STS     &   $36.1\pm3.2$   &   $99.7\pm0.0$   \\
        &   Rand S  &   $95.2\pm1.5$   &   $94.7\pm2.4$   \\
    
    \end{tabular}
    \caption{
    Results of the STS algorithm using the ground truth annotations as predictions. From left to right, the metrics represent: the proportion of images explored relative to the total survey size (lower percentage is preferable), and the detection rate of whales (higher percentage is preferable). Rand S stands for Random Search.
    }
    \label{tab:st_gt}
\end{table}

STS is sensitive to the quality of the model predictions. Thus, it was tested using the ground truth whale annotations. Table \ref{tab:st_gt} presents the average metrics for 20 runs performed on each survey, for the random search baseline and STS. Notice that to find $95\%$ of the populated images the STS requires exploring minimum $13.8\%$ and maximum $47.4\%$ of all images in the survey, while the random search is more exhaustive and inefficient. Even analyzing a significantly minor number of images, STS can find over $95\%$ of the whales, which means it finds the most populated images.

\begin{figure*}[!htb]
    \centering
    \begin{tabular}{  c  c   c   c  c  }
        &   DFOW14  &   DFOW15  &   DFOW16  &   DFOW17  \\
        &   \multicolumn{4}{c}{\underline{Geographical location of all images}}\\
        \multirow{2}{*}{\rotatebox{90}{\small Latitude}}
         &  \includegraphics[trim={1.4cm 1.26cm 0cm 0cm}, clip, height=3cm]{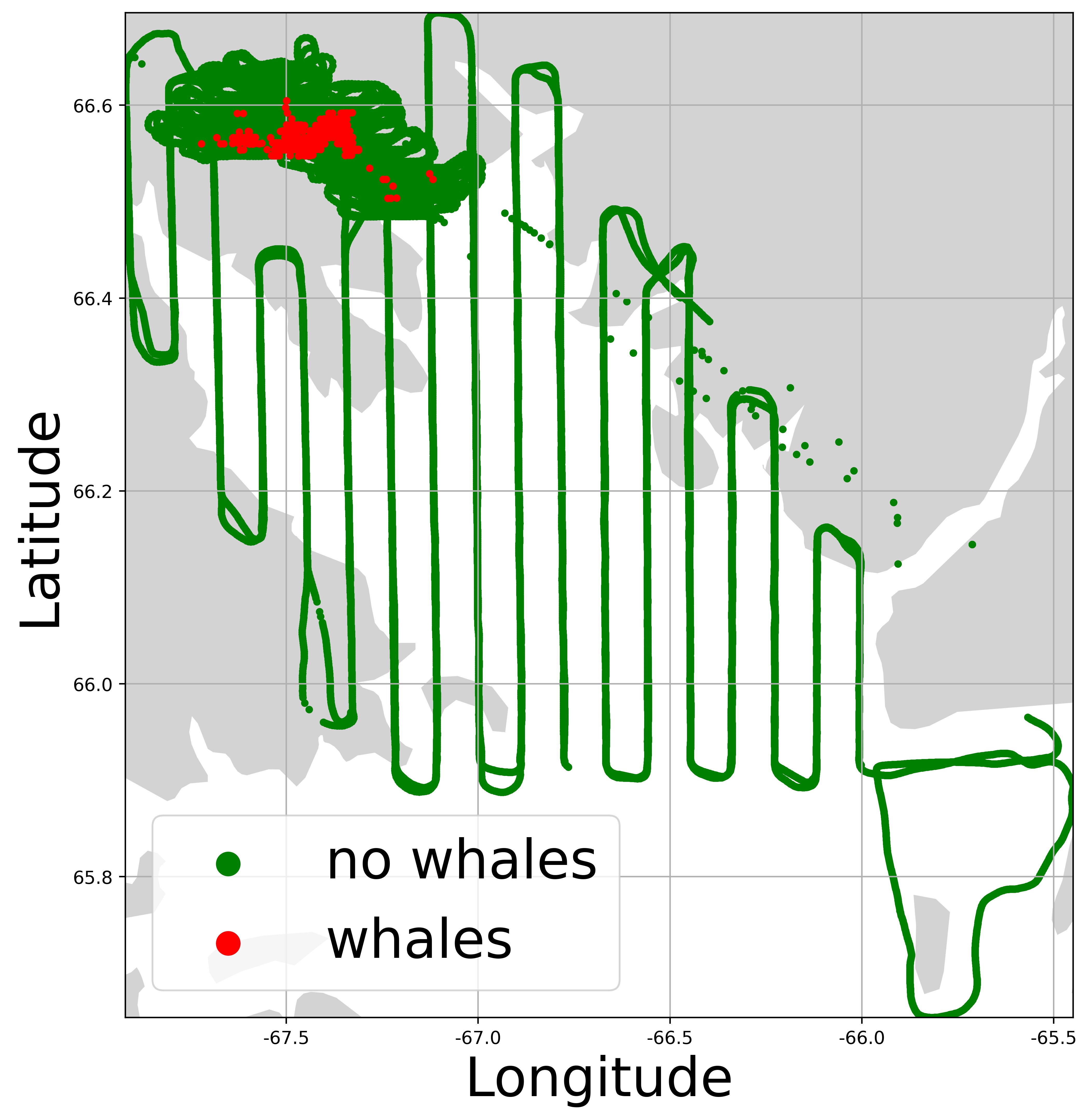} & 
         \includegraphics[trim={1.4cm 1.3cm 0cm 0cm}, clip, height=3cm]{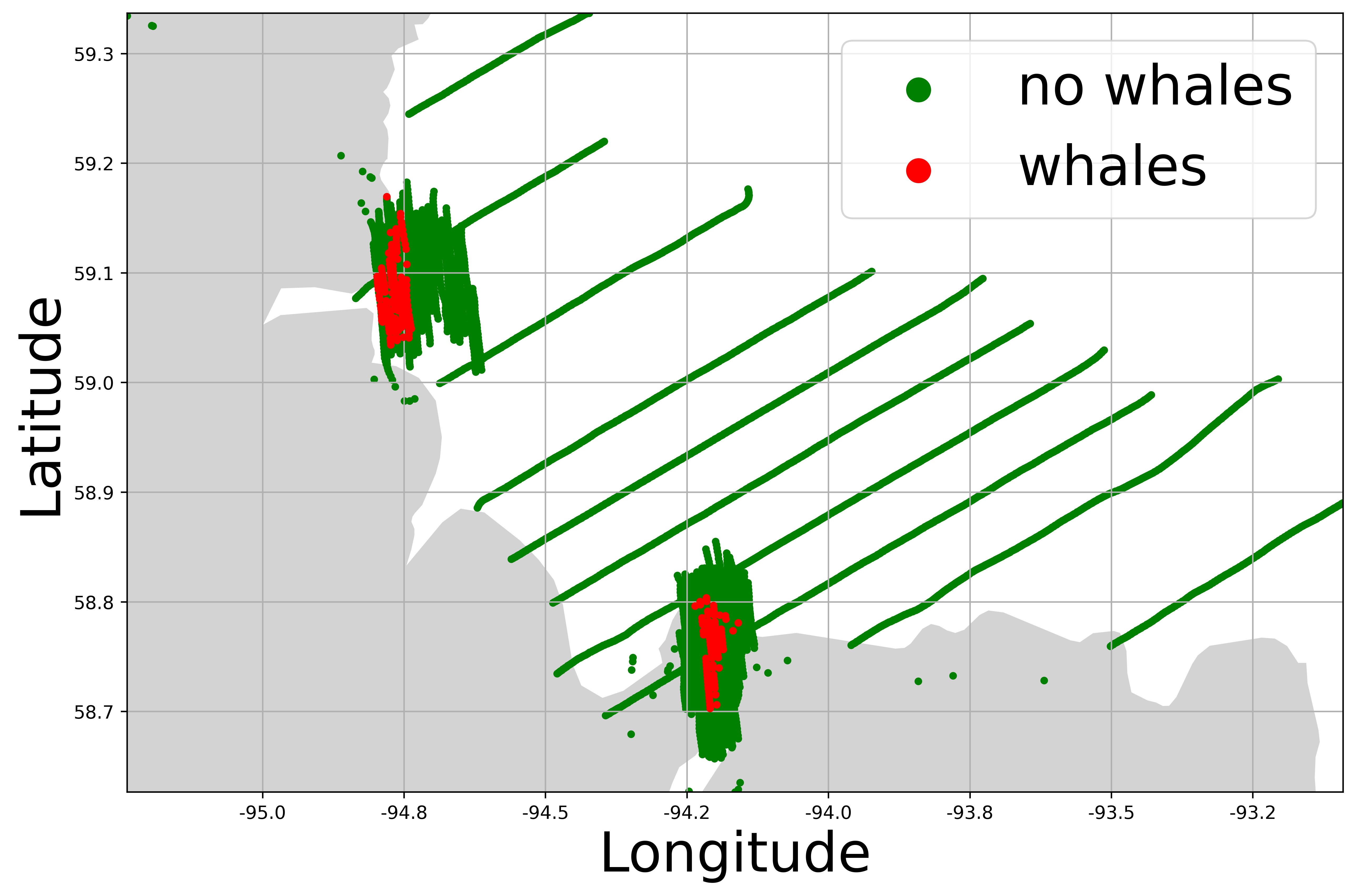} & 
         \includegraphics[trim={1.4cm 1.3cm 0cm 0cm}, clip, height=3cm]{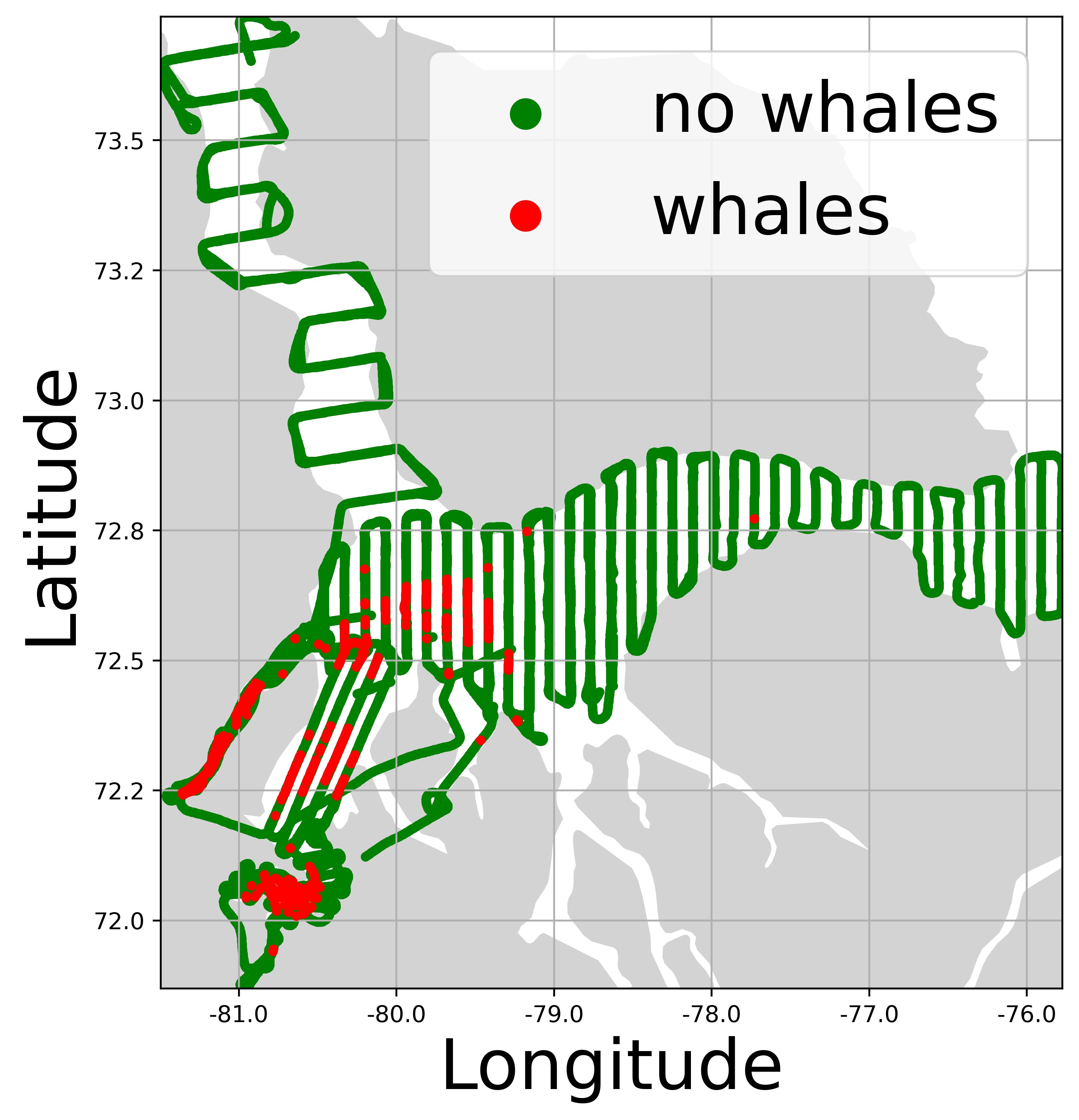} & 
         \includegraphics[trim={1.4cm 1.3cm 0cm 0cm}, clip, height=3cm]{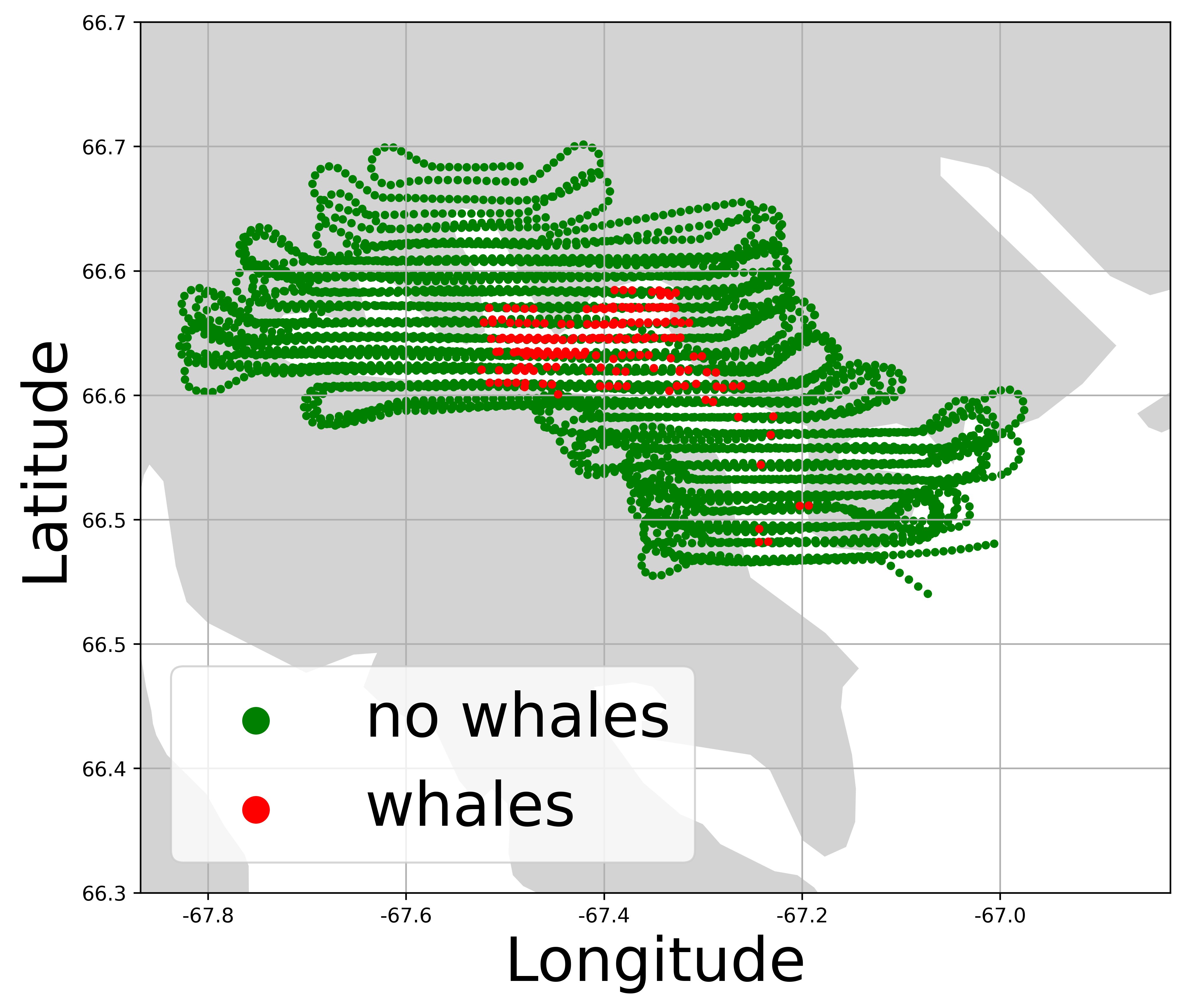} \\
         
        &   \multicolumn{4}{c}{\underline{Geographical location of explored images}}\\
         &  \includegraphics[trim={1.4cm 1.26cm 0cm 0cm}, clip, height=3cm]{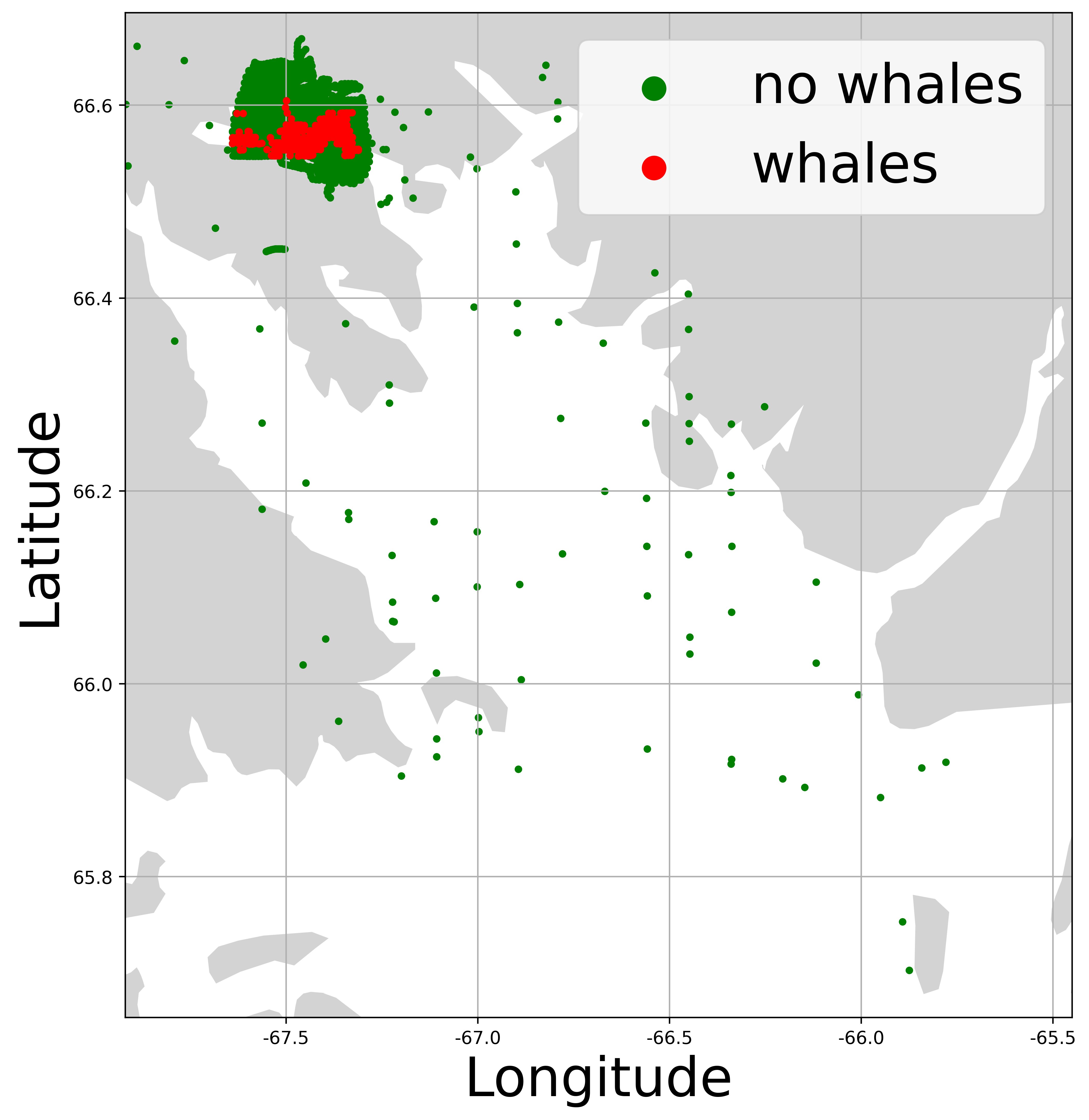} & 
         \includegraphics[trim={1.4cm 1.3cm 0cm 0cm}, clip, height=3cm]{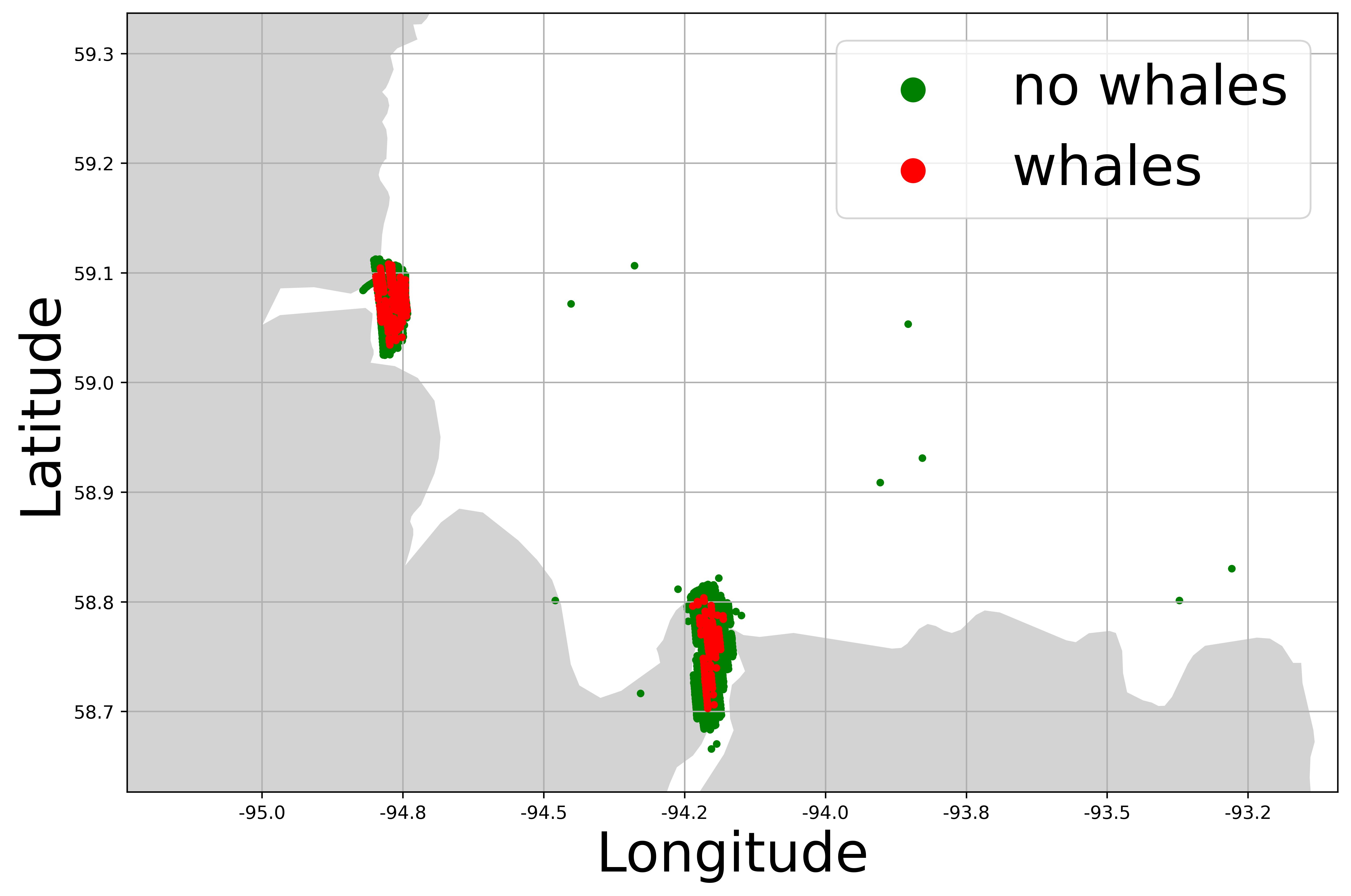} & 
         \includegraphics[trim={1.4cm 1.3cm 0cm 0cm}, clip, height=3cm]{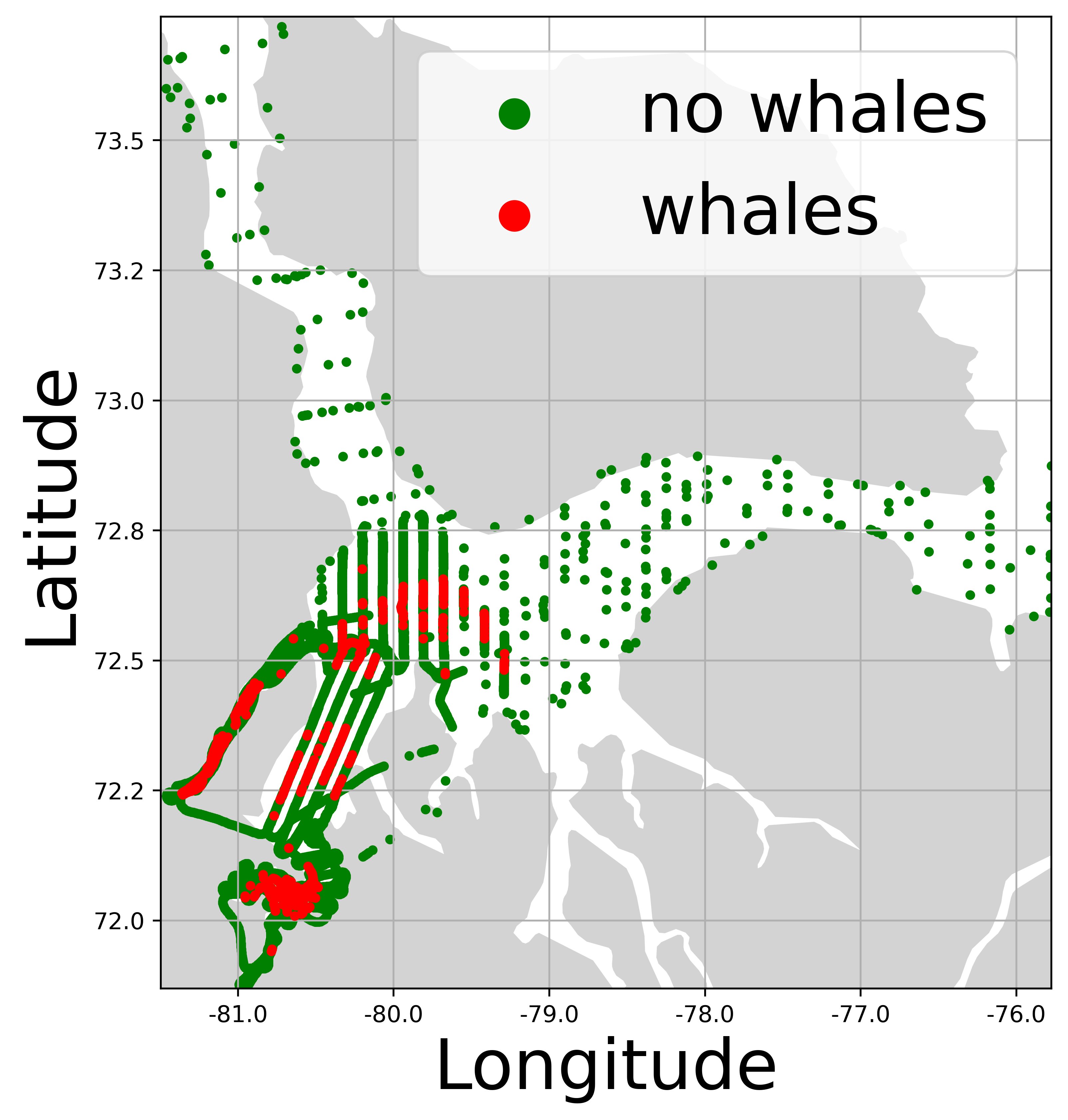} & 
         \includegraphics[trim={1.4cm 1.3cm 0cm 0cm}, clip, height=3cm]{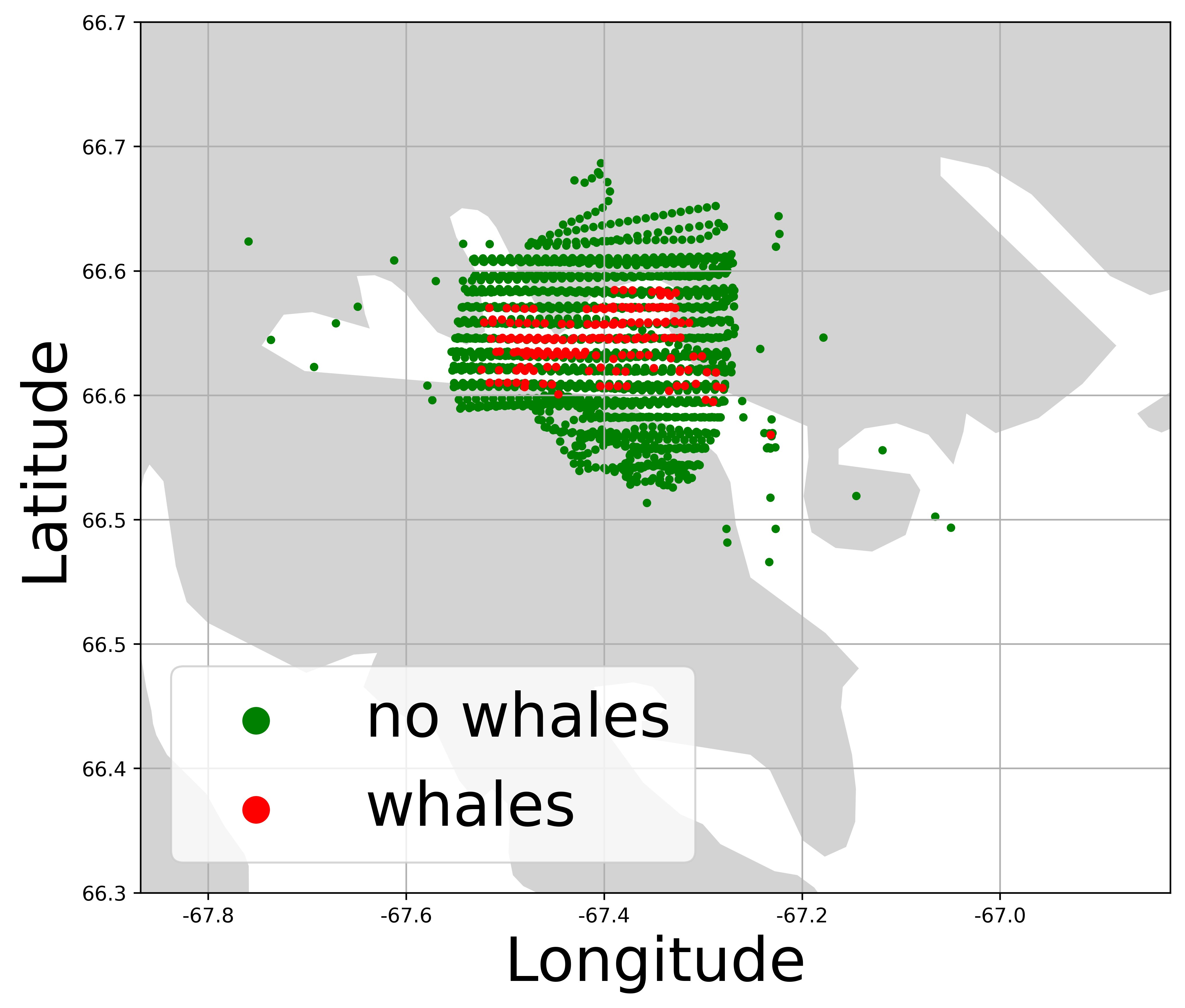} \\
         &  \multicolumn{4}{c}{\small Longitude} \\
    \end{tabular}
    \caption{Top row: Location of all images in the DFO whale surveys. Red dots indicate images with at least one annotated whale, green dots indicate images with no whales annotated. Note the clustering of whale-positive images. Bottom row: Location of the images explored by the STS algorithm. Observe the reduction in images searched.}
    \label{fig:sts_gt}
\end{figure*}

\Cref{fig:sts_gt} illustrates the location of the images in the different surveys, and the images discovered using STS. In the most challenging survey, DFOW16, only $47.4 \pm 1.4\%$ of the total number of images are analyzed, finding $95.2\% \pm 5.1$ of whales. The difficulty with the DFOW16 survey arises because it has several small groups of images containing whales. In contrast, the DFOW14 survey features more densely grouped whales, enabling the search algorithm to cover only $13.8 \pm 1.3\%$ of the images. For DFOW15 whale groups are well defined, and no more images containing whales are discovered because the algorithm stops once it reaches $95\%$. For DFOW17, the group of whales enables the location of most whale-positive images. However, the analysis rate of $36.1 \pm 3.2\%$ reflects the survey size, which is not significantly larger than the count of images containing whales. Overall, the goal of minimizing the number of images to analyze is achieved.

To simulate a more realistic scenario, we evaluated the STS on the DFOW16 survey, this time using predictions from the finetuned OW model, with keywords as caption. Since models designed to detect small targets often produce a high number of false positives, only images with more than 5 predicted instances are considered as whale-containing images, reducing noise in the search space. STS is continuously run 5 times to avoid convergence into a single local group of populated images. As a result, STS explored $58.6\pm8.7 \%$ of all images in the survey, found $39.0\pm1.0 \%$ of the images containing whales, which represented $80.8\pm1.2 \%$ of whales detected. As expected, the percentage of whales decreased compared to the result obtained using the ground truth (see Table \ref{tab:st_gt}), as prediction errors alter the geographical distribution of whales, thereby misrepresenting the expected target grouping. Nonetheless, with the finetuned OW Keyword model, the number of images required to find approximately $80\%$ of the whales was significantly low.

\section{Conclusion}
\label{sec:conclusion}

This paper presents OpenWildlife (OW), an open-vocabulary wildlife detector designed for multi-species detection across diverse geographies. OW leverages a novel application of the GroundingDINO framework, enabling accurate detection of novel species specified by natural language inputs. Combined with our novel Social Target Search (STS) algorithm, organizations can strategically acquire high-resolution satellite imagery to maximize the detection of social species. Given the STS's rapid identification of target-rich images, future work will explore its potential as an acquisition function for active learning. Additionally, we aim to enhance OW's performance by generating more aligned captions using Visual Large Language Models (VLLMs). In summary, by improving both detection and search efficiency, OW holds a strong potential to streamline wildlife monitoring, especially in data-scarce regions. Future developments will also expand OW’s applications and collaboration with conservation agencies globally.
\clearpage
{
    \small
    \bibliographystyle{ieeenat_fullname}
    \bibliography{main}
}

\clearpage 
\setcounter{page}{1} 
\renewcommand{\thesection}{S\arabic{section}} 

\setcounter{section}{0} 
\maketitlesupplementary 




\section{Dataset Descriptions} 
\label{sec:datasets}

The OpenWildlife (OW) model was fine-tuned using the training sets from the datasets listed in Table \ref{tab:training_datasets}, which includes 15 datasets used for open-set continuous fine-tuning. For evaluation, the testing sets from these datasets were specifically reserved to assess the model’s performance under open-set fine-tuning conditions.

Table \ref{tab:novel_species_datasets} provides an overview of the 7 datasets employed for zero-shot detection tasks involving novel species. For these evaluations, the OW model was applied directly to the test images without additional fine-tuning, unless otherwise stated.

Additionally, Table \ref{tab:novel_env_datasets} outlines the 3 datasets used for testing the model’s performance on detecting known species in novel environments. These datasets were chosen to evaluate the model’s ability to transfer knowledge across different geographical regions and environmental conditions.

\begin{table*}[]
    \centering
    \resizebox{\linewidth}{!}{
        \begin{tabular}{l | p{2cm}<{\raggedright} | p{2.5cm}<{\raggedright} | p{1.5cm}<{\raggedright} | p{2cm}<{\raggedright} | p{1.5cm}<{\raggedright} | p{2cm}<{\raggedright}}
            \textbf{Dataset Name} & \textbf{Labels} & \textbf{Region} & \textbf{\# Train Images} & \textbf{\# Train Annotations} & \textbf{\# Test Images} & \textbf{\# Test Annotations} \\ 
            \hline
            \rowcolor{gray!10}
            AED \cite{Naude2019TheDetection.} & Elephant & South Africa, Botswana, Zambia & 1,635 & 12,611 & 439 & 2,970 \\ 
            DFOW14 \cite{MarcouxEstimateSurvey} & Beluga whale & Cumberland Sound, Canada & 476 & 1,935 & 244 & 507 \\ 
            \rowcolor{gray!10}
            DFOW15 \cite{Matthews2017CanadianSurvey} & Beluga whale & West Hudson Bay, Canada & 278 & 3,717 & 186 & 8,000 \\ 
            DFOW17 \cite{Watt2021AbundancePopulation} & Beluga whale & Cumberland Sound Bay, Nunavut, Canada & 666 & 5,181 & 156 & 608 \\ 
            \rowcolor{gray!10}
            IndOcean \cite{Weinstein2022AImagery} & Bird & Pulu Keeling Island and Christmas Island, Oceania & 1,898 & 9,845 & 89 & 326 \\ 
            Izembek \cite{Weiser2023OptimizingCounting} & Brant, Canada goose, Emperor goose & Izembek Lagoon, Alaska, USA & 7,000 & 428,000 & 1,892 & 92,512 \\ 
            \rowcolor{gray!10}
            NewMex \cite{Weinstein2022AImagery} & Bird & New Mexico, USA & 185 & 4,333 & 31 & 282 \\ 
            Palmyra \cite{Weinstein2022AImagery} & Bird & Palmyra Atoll, South Pacific & 298 & 1,315 & 154 & 454 \\ 
            \rowcolor{gray!10}
            Pfeifer \cite{Weinstein2022AImagery} & Penguin & Nelson and King George Islands, Antarctica & 843 & 43,000 & 48 & 2,687 \\ 
            Qian \cite{Qian2023CountingModel} & Penguin & Antarctic Peninsula and South Shetland Islands & 605 & 104,000 & 133 & 30,408 \\ 
            \rowcolor{gray!10}
            SAVMAP \cite{Rey2017DetectingCrowds} & Animal & Kuzikus Wildlife Reserve, Namibia & 523 & 6,337 & 131 & 1,137 \\ 
            Sea Lion \cite{Kaggle2017NOAAKaggle} & Sea lion & Western Aleutian Islands, USA & 579 & 50,000 & 179 & 17,121 \\ 
            \rowcolor{gray!10}
            Seabird \cite{Weinstein2022AImagery} & Bird & North Atlantic & 6,269 & 124,000 & 124 & 7,137 \\ 
            Turtle \cite{Gray2019AImagery} & Turtle & Ostional, Costa Rica & 847 & 1,673 & 212 & 488 \\ 
            \rowcolor{gray!10} 
            Virunga-Gar. \cite{Delplanque2022MultispeciesNetworks} & Alcelaphinae, Buffalo, Kob, Warthog, Waterbuck, Elephant & Democratic Republic of Congo & 928 & 6,962 & 258 & 2,299 \\ 
            \hline
            \textbf{Total} & & & \textbf{23,408} & \textbf{810,492} & \textbf{4,276} & \textbf{166,936} \\ 
        \end{tabular}
    }
    \caption{The 15 datasets used for open-set continuous fine-tuning.}
    \label{tab:training_datasets}
\end{table*}

\begin{table*}[]
    \centering
    \begin{tabular}{l | p{4cm}<{\raggedright} | p{2.5cm}<{\raggedright} | p{1.5cm}<{\raggedright} | p{2cm}<{\raggedright} | p{1.5cm}<{\raggedright} | p{2cm}<{\raggedright}}
        \textbf{Dataset Name} & \textbf{Labels} & \textbf{Region} & \textbf{\# Train Images} & \textbf{\# Train Annotations} & \textbf{\# Test Images} & \textbf{\# Test Annotations} \\ 
        \hline
        \rowcolor{gray!10}
        ArctSeal \cite{EsriAnalyticsTeam2022ArcticHub} & Seal & Alaska, USA & 3,290 & 11,000 & 823 & 2,941 \\ 
        DFOW16 \cite{Marcoux2019CanadianSurveyb} & Narwhal & Eclipse Sound, Canada & 233 & 47,459 & 441 & 948 \\ 
        \rowcolor{gray!10}
        Han \cite{Han2019LivestockNetwork} & Yak, Sheep & Asia & 61 & 3,685 & 17 & 665 \\ 
        Michigan \cite{Weinstein2022AImagery} & Gull & Lake Michigan, USA & 4,850 & 40,000 & 647 & 6,198 \\ 
        \rowcolor{gray!10}
        Polar Bear \cite{Chabot2019MeasuringSpace} & Polar bear & Churchill, Manitoba, Canada & 5 & 7 & 14 & 21 \\ 
        WAID \cite{Mou2023WAID:Drones} & Sheep, Cattle, Seal, Camelus, Kiang, Zebra & Various Locations, Online & 10,000 & 163,000 & 1,437 & 23,820 \\ 
        \rowcolor{gray!10}
        WAfrica \cite{Kellenberger202121000Learning} & Royal tern, Caspian tern, Slender-billed gull, Gray headed gull, Great cormorant, Great white pelican & West Africa & 167 & 18,000 & 43 & 2,637 \\ 
        \hline
        \textbf{Total} & & & \textbf{18,606} & \textbf{283,151} & \textbf{3,422} & \textbf{37,230} \\ 
    \end{tabular}
    \caption{The 7 datasets used for testing on novel species.}
    \label{tab:novel_species_datasets}
\end{table*}

\begin{table*}[]
    \centering
    \begin{tabular}{l | p{4cm}<{\raggedright} | p{2.5cm}<{\raggedright} | p{1.5cm}<{\raggedright} | p{2cm}<{\raggedright} | p{1.5cm}<{\raggedright} | p{2cm}<{\raggedright}}
        \textbf{Dataset Name} & \textbf{Labels} & \textbf{Region} & \textbf{\# Train Images} & \textbf{\# Train Annotations} & \textbf{\# Test Images} & \textbf{\# Test Annotations} \\ 
        \hline
        DFOW23 & Narwhal, Beluga whale, Bowhead whale & North America & 143 & 288 & 343 & 699 \\ 
        Kenya \cite{Eikelboom2019ImprovingDetection} & Zebra, Elephant, Giraffe & Kenya & 393 & 3,000 & 112 & 850 \\ 
        \rowcolor{gray!10}
        Penguins \cite{Liu2020TowardsImagery} & Penguins & Antarctic Peninsula, Antarctica & 172 & 2,019 & 63 & 1,504 \\ 
        \hline
        \textbf{Total} & & & \textbf{708} & \textbf{5,307} & \textbf{518} & \textbf{3,053} \\ 
    \end{tabular}
    \caption{The 3 datasets used for testing on novel environments.}
    \label{tab:novel_env_datasets}
\end{table*}

\section{Hyperparameters}
\label{sec:hyperparameters}

\begin{table*}[]
    \centering
    \begin{tabular}{@{}l|p{6cm}@{}}
        \textbf{Hyperparameter} & \textbf{Value} \\ 
        \hline
        \rowcolor{gray!10} Optimizer & AdamW \\ 
        Learning Rate (lr) & 4e-05 \\ 
        \rowcolor{gray!10} Augmentations & Affine, Brightness, Contrast, RGBShift, HSVShift, JPEG compression, Blur \\ 
        Warm-up Steps & 1000 \\ 
        \rowcolor{gray!10} Max Epochs & 20 \\ 
        \hline
    \end{tabular}
    \caption{Common hyperparameters across all OW models}
    \label{tab:Hyperparameters}
\end{table*}

The hyperparamters used for training all our models are described in Table. \ref{tab:Hyperparameters}.


\section{OW Keyword vs OW Sentence}
\label{sec:comparison_ow_sentence_keyword}

Table \ref{tab:dataset_captions} highlights the differences between the keyword-based and sentence-based captions used for training the OW Keyword and OW Sentence models. Table \ref{tab:caption_vs_no_caption} provides the quantitative comparison of these two approaches on the test portions of fine-tuning dataset.

\begin{table*}[!h]
    \centering
    \begin{tabular}{l | p{2cm}<{\raggedright} | p{12cm}<{\raggedright}}
        \textbf{Dataset} & \multicolumn{2}{c}{\textbf{Caption Type}} \\ 
        \cline{2-3}
        & \textbf{Keyword} & \textbf{Sentence} \\ 
        \hline
        \rowcolor{gray!10}
        DFOW14 & Beluga whale & Species that may be present: beluga whales. Belugas are characterized by their predominantly white coloration, which can appear almost ivory in bright lighting. They have rounded heads and a lack of dorsal fin, giving them a distinctive streamlined shape. In water, they may exhibit a subtle bluish or grayish hue, especially in deeper areas. Their bodies feature smooth, gently curving contours, and they may display light markings that are more prominent on their backs. \\
        AED \cite{Eikelboom2019ImprovingDetection}& Elephant  & Species that may be present: elephant. The species may be large, with a grayish-brown skin that appears rough and wrinkled. They have large ears shaped like fans and long trunks. Their tusks can be visible, protruding from their mouths, which are typically pale in color. \\ 
        \rowcolor{gray!10}
        Sea Lion \cite{Kaggle2017NOAAKaggle} & Sea lion & Species that may be present: sea lion. The species may have a thick, golden to light brown fur that appears lighter in color on their underbelly. They are characterized by a robust body and a pronounced mane, which may appear to be darker compared to the rest of their fur. The animals often have large, rounded foreflippers and a noticeable snout that gives them a distinctive profile.
    \end{tabular}
    \caption{Sample keyword and sentence captions for one image in each dataset.}
    \label{tab:dataset_captions}
\end{table*}

\section{Error Analysis}
\label{sec:error_analysis}

\begin{figure}[ht]
    \centering
    \includegraphics[width=0.5\linewidth]{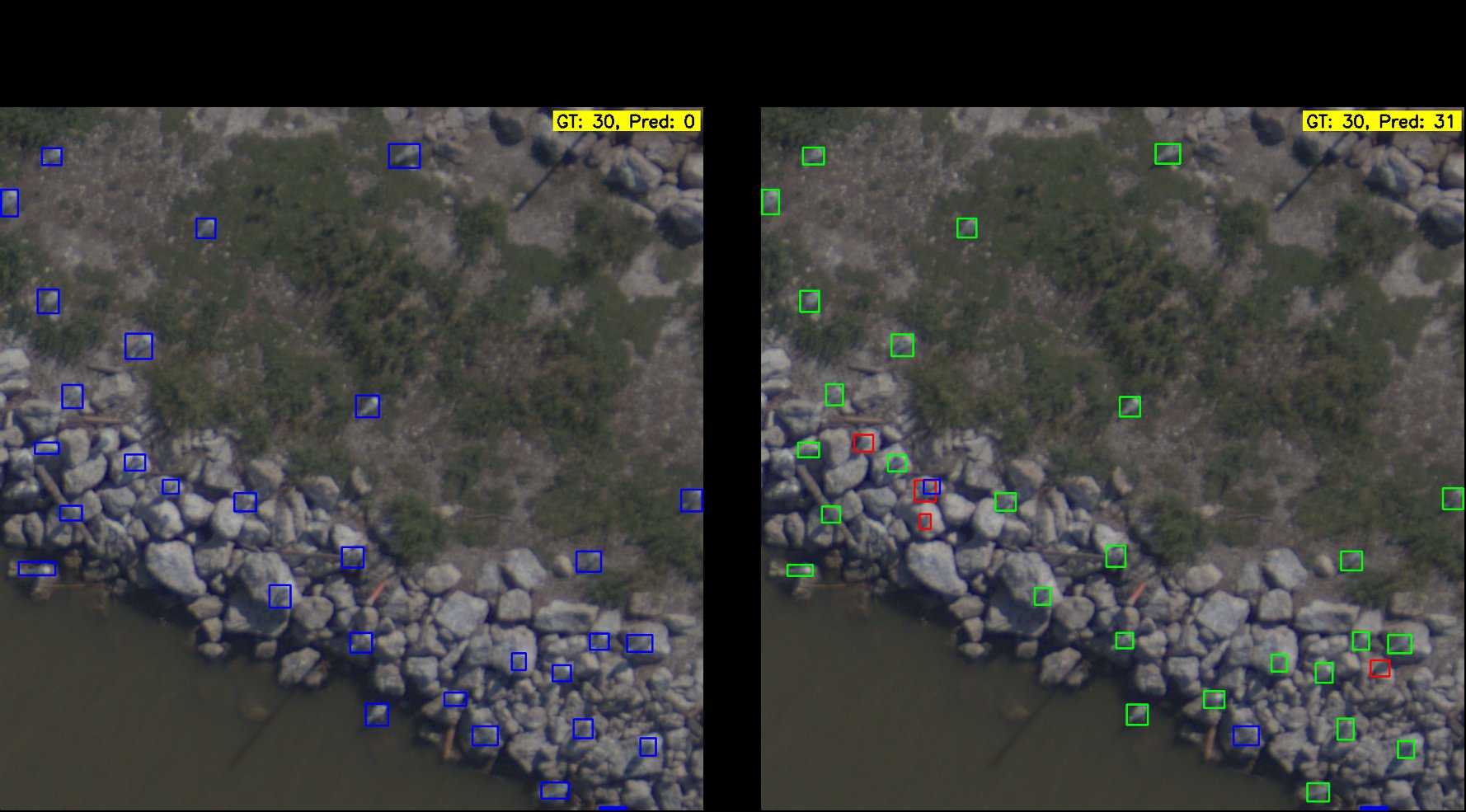}
    \caption{Gull birds blended in rocky terrain. Before fine-tuning OW Keyword couldn't predict any gull; however, after fine-tuning this problem is alleviated}
    \label{fig:gull_birds}
\end{figure}

\begin{figure*}[ht]
    \centering
    \includegraphics[width=0.4\linewidth]{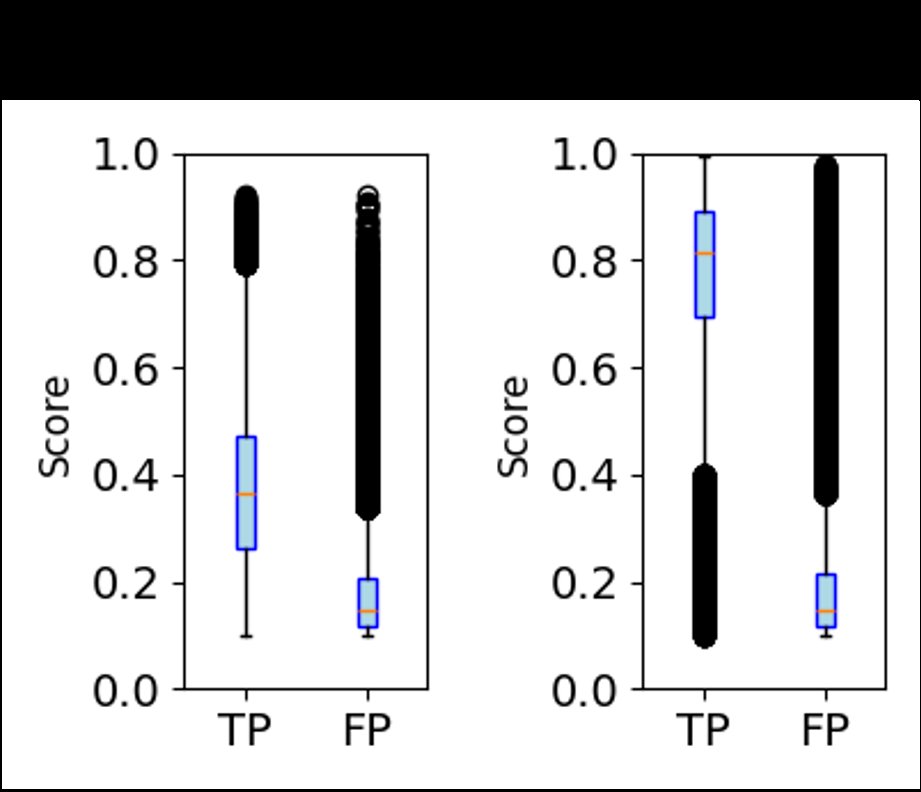}
    \caption{Distribution of OW Keyword model's confidence score on novel species before and after the fine-tuning. Novel species detected by OW can have low confidence scores.}
    \label{fig:confidence_score}
\end{figure*}

To gain deeper insights into the OW model's performance on novel species, we report the true positive (TP), false positive (FP), and false negative (FN) counts across different datasets in Table \ref{tab:confusion_matrix}. First, the number of false positives can be considerably high, likely due to confusion with background elements, as illustrated in Fig. \ref{fig:ow_limitations} in the main paper. Second issue is the subtle signatures of the animals. This issue is present in datasets such as the Michigan, where birds are subtle and challenging to detect (refer Fig. \ref{fig:gull_birds}). Fig. \ref{fig:confidence_score} further shows the distribution of confidence scores before and after fine-tuning. Both the confusion matrix and confidence score distributions underscore the positive impact of fine-tuning.

\begin{table*}
    \centering
    \begin{tabular}{@{} l | p{1.2cm}<{\raggedright} p{1.2cm}<{\raggedright} p{1.2cm}<{\raggedright} | p{1.2cm}<{\raggedright} p{1.2cm}<{\raggedright} p{1.2cm}<{\raggedright} @{}}
        \multirow{2}{*}{\textbf{Dataset}} & \multicolumn{3}{c|}{\textbf{Before finetuning}} & \multicolumn{3}{c}{\textbf{After finetuning}} \\ 
        & \textbf{TP} & \textbf{FP} & \textbf{FN} & \textbf{TP} & \textbf{FP} & \textbf{FN} \\ 
        \hline
        \rowcolor{gray!10}
        WAfrica & 1697 & 2800 & 603 & 2160 & 1295 & 140 \\ 
        Michigan & 1350 & 3841 & 4848 & 5496 & 6250 & 702 \\ 
        \rowcolor{gray!10}
        DFOW16 & 735 & 7454 & 213 & 847 & 7257 & 101 \\ 
        ArctSeal & 2087 & 93632 & 854 & 2921 & 18954 & 20 \\ 
        \rowcolor{gray!10}
        PolarBear & 11 & 408 & 10 & 18 & 28 & 3 \\ 
        Han & 264 & 1640 & 401 & 663 & 961 & 2 \\ 
        \rowcolor{gray!10}
        WAID & 15233 & 110493 & 8587 & 23483 & 13004 & 337 \\ 
    \end{tabular}
    \caption{Confusion matrix for OW keyword on novel species datasets at a prediction score threshold of 0.1}
    \label{tab:confusion_matrix}
\end{table*}

\section{Qualitative Results} \label{sec:qualitative_results}

\begin{figure*}[ht]
    \centering
    \includegraphics[width=\linewidth]{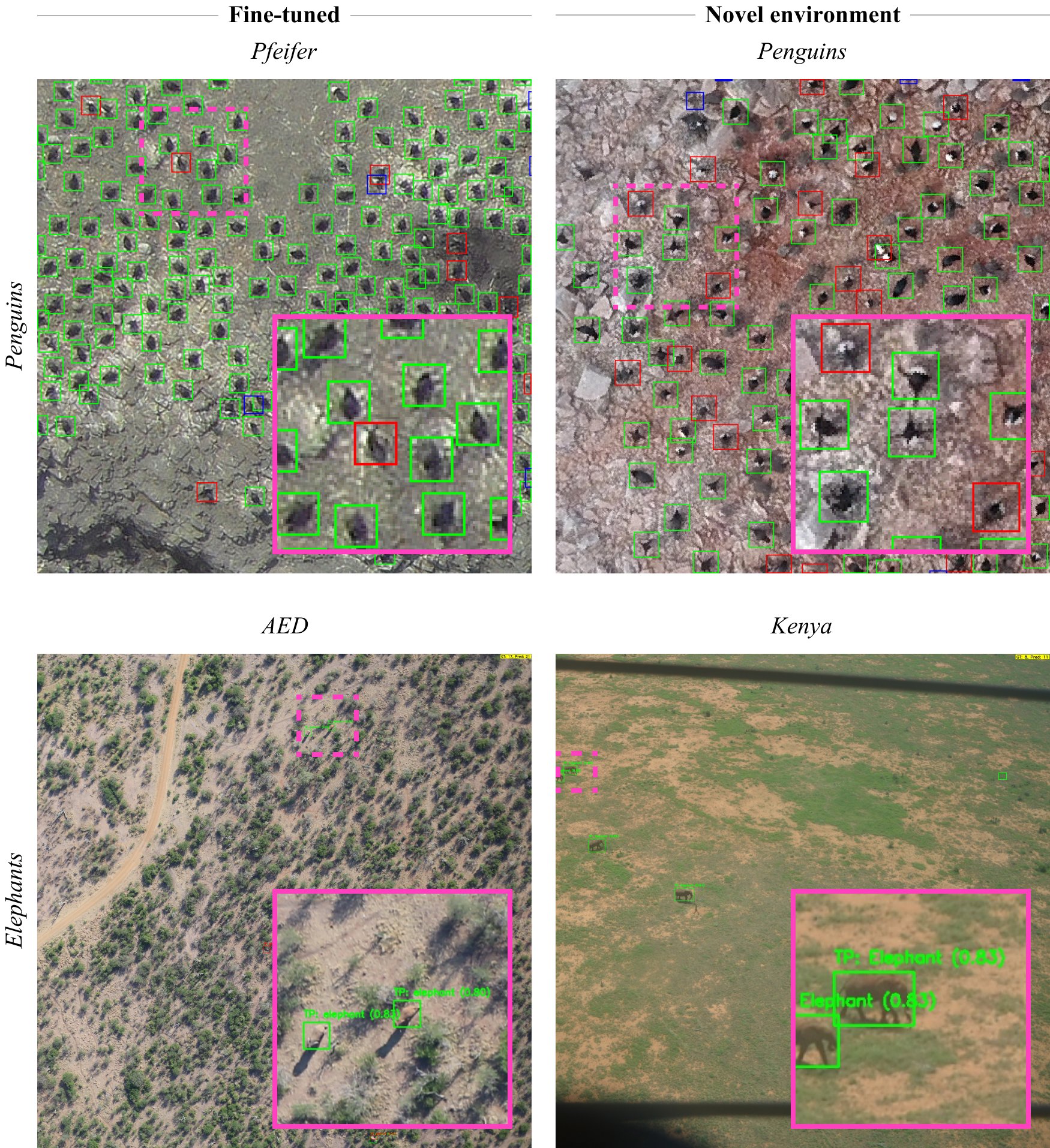}
    \caption{Qualitative results of known species detection in novel environments, demonstrating successful domain transfer for both penguins and elephants. TP is marked as green, FP as red, and FN as blue.}
    \label{fig:novel_environments}
\end{figure*}

\begin{figure*}[ht]
    \centering
    \includegraphics[width=\linewidth]{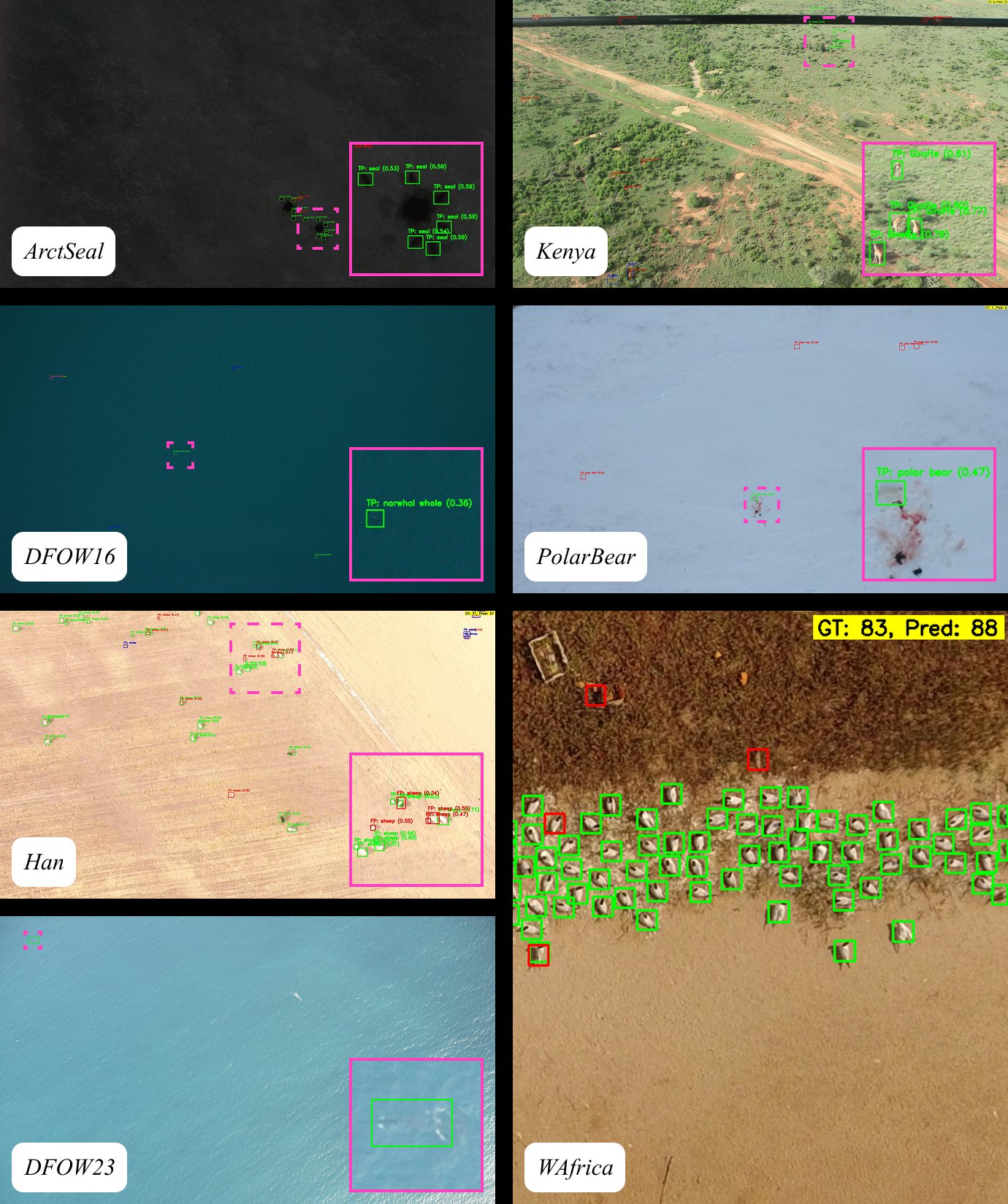}
    \caption{Qualitative results of novel species detection. TP is marked as green, FP as red, and FN as blue.}
    \label{fig:novel_species_full_page}
\end{figure*}

Additional qualitative results of detection on novel species and on known species in novel environments are visualized in Fig. \ref{fig:novel_environments} and Fig. \ref{fig:novel_species_full_page}. These figures demonstrate the OW model’s ability to generalize across diverse environments and species without specific training on the target datasets.

\section*{Note}

All model predictions in this paper are presented in high resolution, with corresponding GIFs showcasing the STS step-wise search region available in the supplementary material. We encourage viewing these resources for improved clarity and visualization.


\begin{table*}[]
    \centering
    \begin{tabular}{@{} l  |r  |r @{}}
        \textbf{Dataset Name} & \textbf{OW Keyword} & \textbf{OW Sentence} \\ 
        \hline
        \rowcolor{gray!10}
        IndOcean & $0.833$ & $\mathbf{0.852}$ \\
        Izembek & $0.343$ & $\mathbf{0.401}$ \\
        \rowcolor{gray!10}
        NewMex & $\mathbf{0.728}$ & $0.642$ \\
        Palmyra & $0.789$ & $\mathbf{0.801}$ \\
        \rowcolor{gray!10}
        Seabird & $\mathbf{0.959}$ & $0.957$ \\
        Pfeifer & $\mathbf{0.842}$ & $0.83$ \\
        \rowcolor{gray!10}
        Qian & $\mathbf{0.671}$ & $0.639$ \\
        Sea lion & $0.662$ & $\mathbf{0.668}$ \\
        \rowcolor{gray!10}
        Turtle \cite{Gray2019AImagery} & $0.26$ & $\mathbf{0.340}$ \\  
        DFOW14 \cite{MarcouxEstimateSurvey} & $\mathbf{0.856}$ & $0.807$ \\ 
        \rowcolor{gray!10}
        DFOW15 \cite{Matthews2017CanadianSurvey} & $\mathbf{0.820}$ & $0.800$ \\
        DFOW17 & $\mathbf{0.857}$ & $0.837$ \\ 
        \rowcolor{gray!10}
        AED \cite{Naude2019TheDetection.} & $0.858$ & $\mathbf{0.864}$ \\
        SAVMAP & $0.379$ & $\mathbf{0.401}$ \\
        \rowcolor{gray!10}
        Viruga-Gar & $0.719$ & $0.719$\\
        \hline
    \end{tabular}
    \caption{Comparison of open-set continuous fine-tuning performance (mAP50) between models trained with keyword and sentence captions. \textbf{Bold values} indicate the better metric for each dataset.}
    \label{tab:caption_vs_no_caption}
\end{table*}



\end{document}